\documentclass[10pt,twocolumn,letterpaper]{article}
\PassOptionsToPackage{table}{xcolor}
\usepackage[T1]{fontenc} %
\usepackage[final]{legacy/cvpr}      %

\usepackage{cuted} 

\usepackage{currfile} 

\usepackage{caption} 

\usepackage{multirow}
\usepackage{adjustbox} 
\usepackage{colortbl}
\definecolor{ExpCText}{RGB}{180,105,25}   
\definecolor{ExpDText}{RGB}{120,70,155}   
\definecolor{tablecolor}{HTML}{EAEAF4}    

\usepackage{tikz}
\usetikzlibrary{arrows.meta,positioning}

\usepackage{microtype}
\usepackage[most]{tcolorbox}
\definecolor{stage0}{HTML}{6C4FB0}   \definecolor{stage0bg}{HTML}{F6F3FB}   
\definecolor{stageA}{HTML}{1659D6}   \definecolor{stageAbg}{HTML}{F1F6FC}   
\definecolor{stageB}{HTML}{D9822B}   \definecolor{stageBbg}{HTML}{FDF6EE}   
\definecolor{coraaccent}{HTML}{5B43A0}   
\definecolor{tablecolor}{HTML}{F1ECFA}   
\definecolor{bestcell}{HTML}{DCD2F2}     
\definecolor{groupband}{HTML}{F2F2F5}    
\definecolor{improve}{HTML}{147A57}      
\definecolor{regress}{HTML}{C8102E}      
\definecolor{famLLM}{HTML}{0E7C86}       
\definecolor{famImp}{HTML}{B03A5B}       
\definecolor{famDiv}{HTML}{2E7D4F}       
\definecolor{famGeo}{HTML}{94700C}       
\newcommand{\famllm}[1]{\textcolor{famLLM}{#1}}
\newcommand{\famimp}[1]{\textcolor{famImp}{#1}}
\newcommand{\famdiv}[1]{\textcolor{famDiv}{#1}}
\newcommand{\famgeo}[1]{\textcolor{famGeo}{#1}}
\newcommand{\famlegend}{Baselines are colored by family: \famllm{\textit{in-LLM}}, \famdiv{\textit{diversity}}, \famimp{\textit{encoder attention}} and \famgeo{\textit{3D-specific}}}
\newcommand{\best}[1]{\cellcolor{bestcell}\textbf{#1}}
\newcommand{\bbmodel}[2]{\rowcolor{groupband}\multicolumn{#1}{@{}l}{\textbf{#2}}}
\newcommand{\dgood}[1]{\textcolor{improve}{$#1$}}
\newcommand{\dneutral}[1]{\textcolor{black}{$#1$}}
\newcommand{\dbad}[1]{\textcolor{regress}{$#1$}}
\newcommand{\dzero}[1]{\textcolor{black!45}{$#1$}}
\newcommand{\stagebadge}[2]{\tikz[baseline=(b.base)]\node[circle,fill=#1,text=white,inner sep=0.6pt,minimum size=1.15em,font=\scriptsize\bfseries](b){#2};}
\newcommand{\Szero}{\textcolor{stage0}{\textsc{Stage1}}}
\newcommand{\SA}{\textcolor{stageA}{\textsc{Stage2}}}
\newcommand{\SB}{\textcolor{stageB}{\textsc{Stage3}}}
\newtcolorbox{insight}[1][Takeaway]{enhanced, breakable, colback=stage0bg, colframe=stage0!45, boxrule=0.7pt, arc=6pt,
  left=2mm, right=2mm, top=1.3mm, bottom=1.3mm, before skip=8pt, after skip=8pt, fontupper=\small,
  before upper={\textbf{\textcolor{stage0}{#1.}}\ }}
\renewcommand{\paragraph}[1]{\vspace{.45em}\noindent\textbf{#1}\hspace{.45em}}

\definecolor{cvprblue}{rgb}{0.21,0.49,0.74}
\usepackage{xr-hyper} %
\usepackage[pagebackref,breaklinks,colorlinks,allcolors=coraaccent]{hyperref}
\hypersetup{hypertexnames=false}
\def\confName{CVPR}
\def\confYear{2027}

\newcommand{\ours}{CORA}
\newcommand{\tablefontsize}{\footnotesize}
\usepackage{pifont}
\usepackage{algorithm}
\usepackage[noend]{algpseudocode}
\algrenewcommand\algorithmiccomment[1]{\hfill{\color{gray}\footnotesize$\triangleright$ #1}}
\newcommand{\cmark}{\textcolor{green!45!black}{\ding{51}}}
\newcommand{\xmark}{\textcolor{red!65!black}{\ding{55}}}
\newcommand{\pmark}{\textcolor{black!50}{$\circ$}}

\makeatletter
\newcommand*{\addFileDependency}[1]{\typeout{(#1)}\@addtofilelist{#1}\IfFileExists{#1}{}{\typeout{No file #1.}}}
\makeatother

\title{CORA: Geometry- and Question-Aware Token Pruning for Multi-View \\3D Reasoning in VLMs}
\author{Nitish Shukla\textsuperscript{1, *} \quad  Nhat-Tan Bui\textsuperscript{2} \quad
 Surgan Jandial\textsuperscript{3, \dag} \quad
 Arun Ross\textsuperscript{1}\\[2pt]
\textsuperscript{1}Michigan State University \quad
\textsuperscript{2}National University of Singapore \quad
\textsuperscript{3}DatologyAI\\[2pt]
\normalsize
\url{https://nitishshukla86.github.io/CORA}
}

\begin{document}
\twocolumn[{%
\renewcommand\twocolumn[1][]{#1}%
\maketitle
\begin{center}
\captionsetup{type=figure}
\includegraphics[width=.99\textwidth]{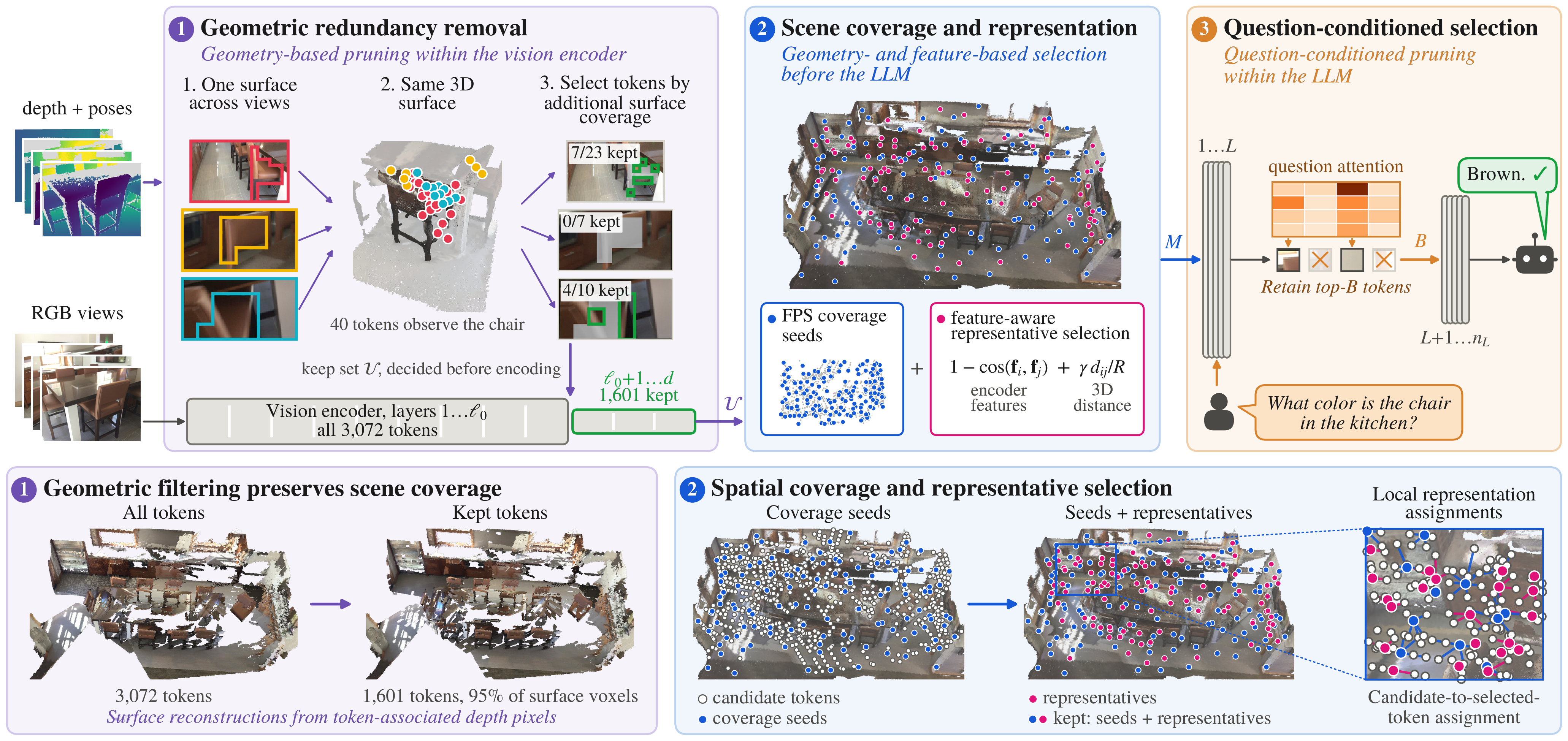}
\vspace{-2.3mm}
\captionof{figure}{\textbf{Overview of \ours{}.} \ours{} coordinates compression to preserve representative scene context for early reasoning and concentrate later computation on question-relevant evidence. (1) Depth and camera poses select tokens covering at least 95\% of observed surface voxels, reducing later vision-encoder computation and funding a larger early LLM context. (2) Spatial coverage and feature-based representative selection populate this context with diverse scene evidence. (3) After layer $L$, question-to-visual attention selects the tokens retained for subsequent layers.}
\label{fig:adv_pipeline}
\end{center}
}]
{\renewcommand{\thefootnote}{*}\footnotetext{Corresponding author: shuklan3@msu.edu.}
\renewcommand{\thefootnote}{\dag}%
\footnotetext{Work done at Carnegie Mellon University.}}

\begin{abstract}
Multi-view 3D question answering requires vision-language models (VLMs) to process thousands of visual tokens from overlapping views. Pruning before joint question--scene processing can discard necessary evidence, while postponing selection increases the cost of early language-model layers. We find that question-to-visual attention provides weak object localization in early layers and becomes more informative in later layers. This motivates CORA, a training-free framework that coordinates compression at three complementary locations: within the vision encoder, before the language model, and within its later layers. First, depth and camera poses identify repeated surface observations, allowing pruning within the vision encoder. The resulting savings fund a larger early language-model context that preserves diverse scene evidence. Second, spatial coverage and representative selection populate this context with tokens capturing distinct scene regions and visual features. Third, after question--scene interaction, attention selects the evidence retained for later layers. The first two stages depend only on the scene and can be reused across questions. Across four VLM backbones and three 3D benchmarks, CORA achieves the highest score in 43 of 48 settings. At reference budgets of 4--4.5\%, it retains 95.7\% of full-token performance at matched encoder-plus-prefill FLOPs, compared with 87.6\% for the strongest baseline. CORA also achieves a $3.58\times$ time-to-first-token speedup under scene reuse and transfers to video question answering.
\end{abstract}

\section{Introduction}
\label{sec:intro}
Understanding a 3D scene requires reasoning about objects, their spatial relationships, and the surrounding layout~\cite{Azuma_2022_CVPR,ma2023sqa3dsituatedquestionanswering}. Recent vision-language models (VLMs) approach this task by processing multiple views of a scene, providing complementary observations for spatial reasoning~\cite{Zhu_2025_ICCV,Zheng_2025_CVPR,qi2025gpt4sceneunderstand3dscenes,yang2025thinkingspacemultimodallarge}. However, this representation introduces substantial redundancy. With 12 views, LLaVA-OneVision-7B receives 8,748 visual tokens~\cite{li2024llavaonevisioneasyvisualtask}, many of which depict the same surfaces from different viewpoints. Processing these overlapping observations increases inference latency and memory consumption, motivating progressive token pruning that preserves question-relevant evidence (\cref{fig:adv_pipeline}).

Existing methods reduce visual tokens either before or within the language model (LLM). Pre-LLM methods select tokens using visual saliency, feature diversity, instruction guidance, or 3D geometry~\cite{Huang_2025_CVPR,bui2026covercoveragebasedtokenpruning,Li_2026_CVPR,Alvar_2025_CVPR,baek20263dzipspatialawarefeaturediversityguided,li2026segprunersemanticgeometricvisualtoken,Zhang_2025_ICCV,NEURIPS2025_2433fec2}. Although these criteria can produce compact scene representations, selection occurs before the language model jointly processes the question and visual evidence, potentially discarding tokens associated with the queried object. In-LLM methods instead use question-to-visual attention to guide selection~\cite{Chen_2024,xing2025pyramiddropacceleratinglargevisionlanguage}, but processing the full visual context in early layers limits their efficiency under tight computational budgets. Methods that prune at multiple locations have also been explored for images and videos~\cite{NEURIPS2025_c573258c,zhang2026vscanrethinkingvisualtoken}; however, they do not explicitly exploit 3D geometry to remove repeated surface observations across views. These limitations motivate a coordinated allocation of computation across visual encoding and language-model depth.

\begin{figure}[t]
\centering
\includegraphics[width=\columnwidth]{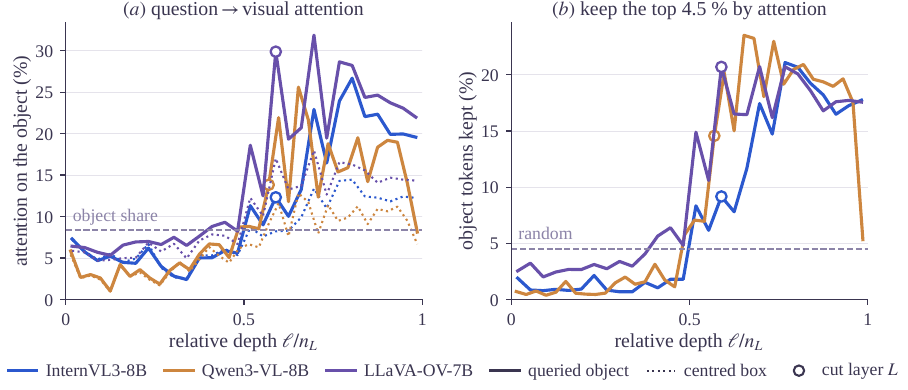}
\vspace{-6mm}
\caption{\textbf{Attention across language-model depth.}
(a) Attention from the last prompt token to the queried object (solid) and an equal-area centered box (dotted); the dashed line indicates the object's fraction of visual tokens.
(b) Fraction of object tokens retained when selecting the top 4.5\% of visual tokens by attention; the dashed line indicates random selection.
Circles mark CORA's pruning layer $L$.}
\label{fig:attn_object}
\vspace{-5mm}
\end{figure}

To guide this allocation, we examine how question-to-visual attention evolves across language-model layers (\cref{fig:attn_object}). In early layers, attention to the queried object is no greater than attention to a centered box of equal area, and attention-based token selection retains fewer object tokens than random selection. In later layers, attention becomes more selective: retaining the top 4.5\% of tokens preserves approximately four times as many object tokens as random selection. Consistent with prior observations on single images~\cite{zhang2026vscanrethinkingvisualtoken,endo2025feather}, these results suggest that pruning should account for the depth at which question-relevant evidence becomes distinguishable. Early layers benefit from diverse scene context, whereas later layers can operate on a compact set of relevant tokens.

We therefore propose CORA, a training-free framework that coordinates token selection across three stages (\cref{fig:adv_pipeline}). First, geometric coverage identifies redundant surface observations using depth and camera poses, allowing their removal from the later vision-encoder layers. The resulting computational savings support a larger visual context in the early language-model layers. Second, spatial coverage and representative selection construct a compact context that preserves diverse scene evidence. Third, after the language model has jointly processed the scene and question, question-to-visual attention selects the tokens retained for subsequent layers. Together, these stages preserve broad evidence before question-conditioned selection becomes reliable and reduce computation thereafter. The first two stages depend only on the scene and can be reused across questions.

We evaluate CORA across four VLM backbones---InternVL3~\cite{zhu2025internvl3exploringadvancedtraining}, two Qwen3-VL variants~\cite{bai2025qwen3vltechnicalreport}, and LLaVA-OneVision~\cite{li2024llavaonevisioneasyvisualtask}---on ScanQA~\cite{Azuma_2022_CVPR}, SQA3D~\cite{ma2023sqa3dsituatedquestionanswering}, and VSI-Bench~\cite{yang2025thinkingspacemultimodallarge}. CORA achieves the best performance in 43 of 48 settings, including all 12 settings at the tightest reference budgets, where its mean improvement over the strongest baseline is 2.53 points. Replacing ground-truth geometry with VGGT predictions~\cite{Wang_2025_CVPR} reduces performance by at most 0.88 points on backbones using encoder-side pruning. CORA further transfers to VideoMME~\cite{Fu_2025_CVPR} and MVBench~\cite{Li_2024_CVPR}, outperforming the evaluated 3D-specific baselines in 15 of 16 settings.

Our contributions are threefold:
\begin{itemize}
    \item We analyze the evolution of question-to-visual attention in multi-view 3D reasoning and show that later layers localize queried objects more reliably than early layers, motivating delayed question-conditioned pruning.
    \item We introduce CORA, a training-free framework that combines geometric redundancy removal, representative scene selection, and question-conditioned pruning to allocate computation across the vision encoder and language model.
    \item We demonstrate consistent improvements across four VLM backbones and three 3D benchmarks, together with measured inference speedups and transfer to two video benchmarks.
\end{itemize}

\section{Related Work}
\label{sec:related}

\noindent\textbf{Visual token pruning.}
FastV~\cite{Chen_2024} and PyramidDrop~\cite{xing2025pyramiddropacceleratinglargevisionlanguage} select visual tokens using attention scores within the language model.
Before the language model, DivPrune~\cite{Alvar_2025_CVPR} selects tokens to preserve visual-feature diversity, while CDPruner~\cite{NEURIPS2025_2433fec2} additionally conditions selection on the instruction.
VisPruner~\cite{Zhang_2025_ICCV} uses vision-encoder attention to identify important visual tokens.
Multi-stage frameworks also combine compression across model components: VScan pairs pre-LLM reduction with middle-layer pruning~\cite{zhang2026vscanrethinkingvisualtoken}, HoliTom combines outer- and inner-LLM merging~\cite{NEURIPS2025_c573258c}, and METEOR prunes across encoding, fusion, and decoding in multi-encoder models~\cite{liu2025meteor}.
These methods do not explicitly measure the coverage of observed 3D surfaces.
\ours{} combines geometry-based redundancy removal and representative scene selection with  question-conditioned pruning.

\noindent\textbf{Multi-view 3D token pruning.}
CoVeR~\cite{bui2026covercoveragebasedtokenpruning} selects tokens based on spatial coverage, SeGPruner~\cite{li2026segprunersemanticgeometricvisualtoken} combines encoder saliency with geometric diversity, and 3DZip~\cite{baek20263dzipspatialawarefeaturediversityguided} combines voxel aggregation, diverse anchor selection, and merging.
These methods reduce tokens before the LLM using question-independent scene representations.
GeoGR further combines projector-level scene compression with recoverable token routing within the LLM under compression-aware post-training~\cite{zheng2026geogr}.
\ours{} instead targets three computational locations: geometric reduction within the vision encoder, representative selection before the LLM, and question-conditioned selection within the LLM.
\Cref{tab:related} compares the selection criteria and pruning locations of the evaluated methods. Among these methods, only \ours{} combines geometry with question guidance across all three locations, preserving scene evidence for early language processing while reducing computation in later layers.
\begin{table}[t]
\centering
\tablefontsize
\setlength{\tabcolsep}{3pt}
\caption{\textbf{Comparison with baselines.} Geom.: geometry; Enc./Pre./LLM: pruning location; Query: question-conditioned; \pmark: partial support.}
\vspace{-2mm}
\label{tab:related}
\begin{tabular*}{\columnwidth}{@{\extracolsep{\fill}}l ccccc@{}}
\toprule
Method & Geom. & Enc. & Pre. & LLM & Query \\
\midrule
FastV~\cite{Chen_2024} & \xmark & \xmark & \xmark & \cmark & \pmark \\
PyramidDrop~\cite{xing2025pyramiddropacceleratinglargevisionlanguage} & \xmark & \xmark & \xmark & \cmark & \cmark \\
DivPrune~\cite{Alvar_2025_CVPR} & \xmark & \xmark & \cmark & \xmark & \xmark \\
CDPruner~\cite{NEURIPS2025_2433fec2} & \xmark & \xmark & \cmark & \xmark & \cmark \\
VisPruner~\cite{Zhang_2025_ICCV} & \xmark & \xmark & \cmark & \xmark & \xmark \\
SeGPruner~\cite{li2026segprunersemanticgeometricvisualtoken} & \cmark & \xmark & \cmark & \xmark & \xmark \\
CoVeR~\cite{bui2026covercoveragebasedtokenpruning} & \cmark & \xmark & \cmark & \xmark & \xmark \\
3DZip~\cite{baek20263dzipspatialawarefeaturediversityguided} & \cmark & \xmark & \cmark & \xmark & \xmark \\
\midrule
\rowcolor{tablecolor}\textbf{\ours{}} & \cmark & \cmark & \cmark & \cmark & \cmark \\
\bottomrule
\end{tabular*}
\end{table}

\section{Method}
\label{sec:adv_method}

\subsection{Overview}
\label{sec:adv_problem}
\label{sec:adv_overview}

Given RGB-D views with known camera poses and a question $q$, \ours{} reduces the $T$ visual tokens processed by a frozen VLM through three nested selections $[T]\supseteq\mathcal{U}\supseteq S\supseteq S'$ (\cref{fig:adv_pipeline}). Here, $[T]=\{1,\ldots,T\}$ indexes the full visual-token sequence, and $\mathcal{U}$ is the subset retained by geometric pruning.
Stage~1 uses geometry to remove repeated observations of the same surface inside the vision encoder.
Stage~2 selects a representative scene context $S$ for the first $L$ language-model layers, where $L$ denotes the layer after which question-conditioned pruning is applied.
Stage~3 retains the question-relevant subset $S'$ for the remaining layers.
Stages~1 and~2 are question-independent, allowing their selections to be reused across questions about the same scene.
Each selection preserves the retained tokens' features and positions without merging or synthesizing tokens. \Cref{alg:cora} summarizes the three stages.

\subsection{Token and compute budgets}
\label{sec:adv_budget}
We distinguish between matching the layer-averaged visual-token count and matching encoder-plus-prefill computation.  Let $n_L$ be the number of language-model layers and $K$ the reference visual-token count per layer. A pre-LLM pruner retains $K$ of the original $T$ visual tokens in every layer; thus, $K/T$ is its retained-token ratio. In the token-matched setting, \ours{} has the same layer-averaged count $K$, but retains $M$ tokens in layers $1,\ldots,L$ and $B$ tokens in the remaining $n_L-L$ layers. Defining the late-to-early token ratio as $r=B/M$, equating the counts gives $LM+(n_L-L)B=n_LK$, and hence
\begin{equation}
  M = \mathrm{round}\!\left(\frac{n_L K}{L + (n_L-L)r}\right), \quad B = \mathrm{round}(rM).
  \label{eq:adv_budget}
\end{equation}
We set $L=\lfloor16n_L/28\rfloor$, about 57\% of the depth, and $r=1/2$, which gives $M\approx1.28K$.
Thus, the early layers receive approximately 28\% more tokens than with pre-LLM pruning, balanced by a smaller late context at the same layer-averaged budget.

\noindent\textbf{Compute-matched setting.}
Stage~1 reduces encoder computation, allowing a larger early context at the same total computational cost. We measure computation in floating-point operations (FLOPs). Let $E$ denote the full vision encoder's cost and $f\le1$ the fraction of that cost remaining with Stage~1, including the encoder layers before pruning. Let $F(m)$ denote the LLM-prefill cost with $m$ visual tokens in the first $L$ layers and $\mathrm{round}(rm)$ thereafter. The compute-matched counts $M'$ and $B'$ specify the early and late visual contexts, respectively.
We set
\begin{equation}
  \begin{aligned}
    M' &= \max\{m \in \mathbb{N} \mid fE + F(m) \le E + F(M)\}, \\
    B' &= \mathrm{round}(rM').
  \end{aligned}
  \label{eq:adv_compute_budget}
\end{equation}
This selects the largest early context within the reference cost, $E+F(M)$, which combines a full encoder with the token-matched progressive schedule. Reallocating encoder savings can therefore raise the layer-averaged token count above the reference budget. Reported percentages denote the reference ratio $K/T$, rather than CORA's retained fraction at each layer. Without encoder pruning, the compute-matched and token-matched counts coincide. We use the compute-matched allocation below unless stated otherwise.

\begin{algorithm}[t]
\caption{\ours{}.}
\label{alg:cora}
\small
\begin{algorithmic}[1]
\Require posed RGB-D views, question $q$, budget $K$, frozen VLM
\State compute $(M',B')$ with \cref{eq:adv_compute_budget}
\Statex \stagebadge{stage0}{1}\ \textcolor{stage0}{\textbf{Geometric coverage}}
\State $\mathcal{U}\gets\Szero(\text{depth, poses}, M')$
\State run the first half of the encoder on all units, the second half on $\mathcal{U}$
\Statex \stagebadge{stageA}{2}\ \textcolor{stageA}{\textbf{Representative scene selection}}
\State $S\gets\SA(\mathcal{U}, M')$
\Statex \stagebadge{stageB}{3}\ \textcolor{stageB}{\textbf{Question-conditioned selection}}
\State run LLM layers $1,\dots,L$ on $[\,S;q\,]$
\State $S'\gets\SB(S,B')$ \Comment{attention of layer $L{+}1$}
\State run the remaining layers on $S'$ and decode
\end{algorithmic}
\end{algorithm}

\subsection{Stage 1: Geometric redundancy removal}
\label{sec:adv_stage0}

Stage~1 removes repeated surface observations across views using depth and camera poses.
Each unit $u$ is the image region of one language model token.
We back-project its pixels with valid depth and let $\mathcal{V}_u$ be the voxels containing the back-projected points.
Let $\mathcal{V}=\bigcup_u\mathcal{V}_u$ denote all observed surface voxels and $U$ a candidate subset of units. The coverage target $\eta$ specifies the fraction of these voxels that must remain represented. Stage~1 seeks the smallest subset satisfying this target:
\begin{equation}
  \min_{U}\, |U| \quad\text{s.t.}\quad \Bigl|\bigcup_{u\in U}\mathcal{V}_u\Bigr| \ge \eta\,|\mathcal{V}|.
  \label{eq:adv_cover}
\end{equation}
We approximate this partial set-cover problem using lazy greedy selection, repeatedly adding the unit that covers the most previously uncovered voxels.
The retained set $\mathcal{U}$ contains at least enough units to fill Stage~2's early-context token budget.
The encoder processes all units in its first half and only the retained units in its second half, allowing retained representations to incorporate full-image context before pruning. We prune after half of the encoder layers, which keeps accuracy while saving encoder computation.

Stage~1 is applied to InternVL3 and Qwen3-VL, which merge $2\times2$ encoder patches into each language-model token. LLaVA-OneVision instead maps each $14$-pixel patch to a separate token. At this finer granularity, coverage-based selection can remove tokens more aggressively. We therefore preserve its full encoder context and apply only Stages~2--3.

\subsection{Stage 2: Scene coverage and representation}
\label{sec:adv_seeds}
\label{sec:adv_fl}

After geometric reduction, Stage~2 fills the early-context token budget by selecting tokens that preserve spatial coverage and represent the appearance of local scene regions.
Let $\mathcal{C}\subseteq\mathcal{U}$ contain candidate tokens with finite, non-zero 3D positions, excluding protected padding tokens (details in the supplementary material).
Each candidate $i$ has an encoder feature vector $\mathbf{f}_i$ and a back-projected 3D position $\mathbf{x}_i$. Protected padding tokens correspond to image padding retained within the encoder.
If the number of valid candidates does not exceed the early-context token budget, we keep all candidates.

\noindent\textbf{Coverage seeds.}
We first select spatially distributed coverage seeds, denoted by $S_0$, using a fixed fraction of the early-context token budget. We use farthest-point sampling (FPS)~\cite{Eldar_1997,Gonzalez_1985}, which repeatedly selects the candidate farthest from its nearest existing seed, starting from
the candidate nearest to the centroid of the scene.
Let $d_s$ be the distance from the last seed to its nearest earlier seed.
Since FPS distances never increase, every candidate lies within $d_s$ of a seed.
The neighborhood radius is $R=\max(2d_s,0.02\,\mathrm{m})$: twice the final seed spacing, with a 2\,cm lower bound. Candidates assigned to the same nearest seed are then within this radius of one another. For a single seed, the spacing is defined as the maximum candidate-to-seed distance.

\noindent\textbf{Representative selection.}
Let $\mathcal{N}_i$ be the candidates within $R$ of candidate $i$.
For a candidate $i$ and a possible representative $j$, the assignment cost $c_{ij}$ combines cosine distance between token features and Euclidean distance between their 3D positions, normalized by the neighborhood radius. The weight $\gamma\ge0$ controls the spatial contribution, and $\bar c=1+\gamma$ is the cost assigned outside the neighborhood:
\begin{equation}
  c_{ij} =
  \begin{cases}
    1-\cos(\mathbf{f}_i,\mathbf{f}_j) + \gamma\,\lVert\mathbf{x}_i-\mathbf{x}_j\rVert/R, & j\in\mathcal{N}_i,\\
    \bar c = 1+\gamma, & \text{otherwise}.
  \end{cases}
  \label{eq:adv_cost}
\end{equation}
Surfaces seen by many views would dominate an unweighted objective, so we weight each candidate by $w_i$, the inverse of the number of views that observe a candidate near $\mathbf{x}_i$.
We fill the remaining budget by minimizing the weighted facility-location cost, assigning each candidate to its least-cost selected representative:
\begin{equation}
  F(S) = \sum_{i\in\mathcal{C}} w_i \min\Bigl(\bar c,\ \min_{j\in S} c_{ij}\Bigr),
  S_0\subseteq S\subseteq\mathcal{C},\ |S|=M'.
  \label{eq:adv_fl}
\end{equation}
The cap limits the contribution of distant or dissimilar candidates; the constraints preserve coverage seeds and fill the early-context budget. Minimizing this cost is equivalent to maximizing a monotone submodular facility-location gain with nonnegative similarities. For efficient selection, we add high-gain candidates in batches, with half-radius separation to limit local redundancy.
\vspace{-1mm}
\subsection{Stage 3: Question-conditioned selection}
\label{sec:adv_stageb}

Stage~3 uses the contextualized representations after layer $L$ to select question-relevant evidence. Let $\mathcal{Q}$ index the question tokens and $A_{h,q,t}$ denote attention from question token $q\in\mathcal{Q}$ to visual token $t\in S$ in head $h$, computed using the query and key projections of layer $L{+}1$.
With $H$ attention heads, each visual token $t$ receives a peak score $p_t$ and a mean score $a_t$,
\begin{equation}
  p_t = \frac{1}{H}\sum_{h}\max_{q\in\mathcal{Q}} A_{h,q,t}, \quad
  a_t = \frac{1}{H|\mathcal{Q}|}\sum_{h}\sum_{q\in\mathcal{Q}} A_{h,q,t}.
  \label{eq:adv_peakmean}
\end{equation}
The peak score captures strong attention from an individual question token, such as an object name, while the mean score captures attention across the question.
We independently min-max normalize both scores to $[0,1]$ over the early context and sum them to obtain $s_t$. Here, $\mathrm{mm}$ denotes this normalization, mapping equal-valued vectors to zero, and $\mathrm{Top}_{B'}$ selects the highest-scoring tokens within the late-context budget:
\begin{equation}
  s_t=\mathrm{mm}(p)_t+\mathrm{mm}(a)_t,\qquad S'=\mathrm{Top}_{B'}(s).
  \label{eq:adv_score}
\end{equation}
The removed tokens are dropped from the hidden states and the KV cache, and the kept tokens retain their  positions.

\section{Experiments and Results}
\label{sec:exp}
\noindent\textbf{Setup.}
We evaluate on ScanQA~\cite{Azuma_2022_CVPR} validation (4,675 questions), SQA3D~\cite{ma2023sqa3dsituatedquestionanswering} test (3,519 questions), and the ScanNet and ARKitScenes subsets of VSI-Bench~\cite{yang2025thinkingspacemultimodallarge} (3,672 questions).
We report EM@1 on ScanQA and SQA3D and the official score on VSI-Bench.
Each scene uses 12 uniformly sampled RGB-D views with camera poses.
We use frozen InternVL3-8B~\cite{zhu2025internvl3exploringadvancedtraining}, Qwen3-VL-8B and Qwen3-VL-4B~\cite{bai2025qwen3vltechnicalreport}, and LLaVA-OneVision-7B~\cite{li2024llavaonevisioneasyvisualtask} (LLaVA-OV-7B in tables).
We evaluate four reference budgets: 4.5, 9, 14, and 23\% on ScanQA and VSI-Bench, and 4, 8, 17, and 26\% on SQA3D.
These percentages specify the token budget of the reference configuration relative to the full visual-token count.
Unless stated otherwise, \ours{}'s token counts are calibrated to match the encoder-plus-prefill FLOPs of this reference.
LLaVA-OneVision omits Stage~1.
\begin{figure}[t]
\centering
\includegraphics[width=\columnwidth]{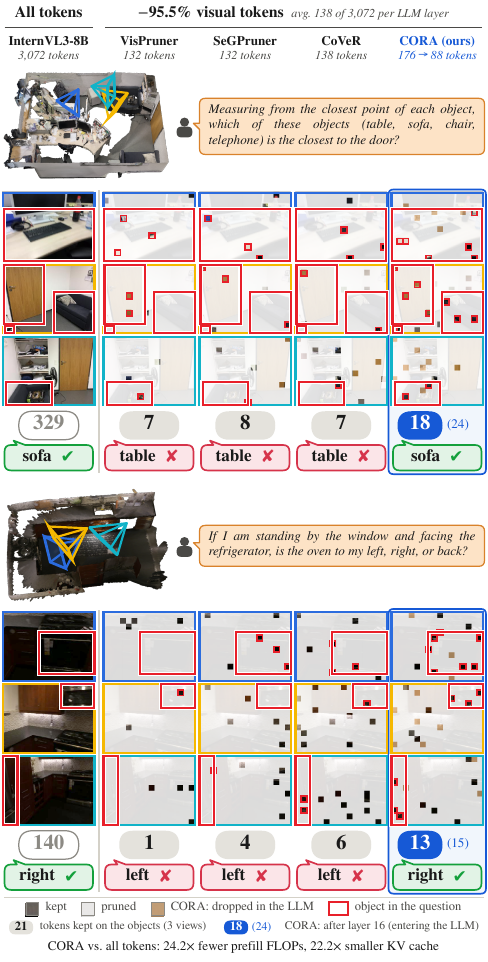}
\vspace{-6mm}
\caption{\textbf{Qualitative comparison on InternVL3-8B at the 4.5\% reference budget.} At a matched token budget, \ours{} preserves more visual tokens associated with the queried objects.}
\label{fig:qual_ivl}
\vspace{-3mm}
\end{figure}

\noindent\textbf{Hyperparameters.}
We use the same hyperparameters across all backbones, budgets, and benchmarks: a geometric coverage target of 95\%, a voxel size of 0.05\,m, a coverage-seed fraction of 0.4, and a spatial-distance weight of 0.5.
Additional implementation settings are provided in the supplementary material.

\noindent\textbf{Baselines.}
We compare with eight training-free pruning methods.
FastV~\cite{Chen_2024} and PyramidDrop~\cite{xing2025pyramiddropacceleratinglargevisionlanguage} use attention scores to prune visual tokens within the language model.
DivPrune~\cite{Alvar_2025_CVPR} and CDPruner~\cite{NEURIPS2025_2433fec2} select diverse visual features before the language model, with CDPruner additionally using the instruction.
VisPruner~\cite{Zhang_2025_ICCV} uses vision-encoder attention for selection.
SeGPruner~\cite{li2026segprunersemanticgeometricvisualtoken}, CoVeR~\cite{bui2026covercoveragebasedtokenpruning}, and 3DZip~\cite{baek20263dzipspatialawarefeaturediversityguided} exploit explicit 3D geometry to reduce redundancy across views.
FastV and PyramidDrop exceed the 4--4.5\% reference budgets within their first two LLM layers and are therefore infeasible.

\begin{table*}[!t]
\centering
\tablefontsize
\setlength{\tabcolsep}{1.9pt}
\renewcommand{\arraystretch}{0.85}
\setlength{\aboverulesep}{1pt}
\setlength{\belowrulesep}{1pt}
\caption{\textbf{Main results at matched encoder-plus-prefill FLOPs.} Shaded: best; underlined: second best; --: budget infeasible. Avg.: mean over all benchmarks and budgets for each backbone, reported only for methods evaluated at every budget. Parentheses indicate the gain over the strongest baseline's mean. \famlegend.}
\label{tab:adv_main}
\begin{tabular*}{\textwidth}{@{\extracolsep{\fill}}l cccc cccc cccc c@{}}
\toprule
 & \multicolumn{4}{c}{ScanQA}
 & \multicolumn{4}{c}{SQA3D}
 & \multicolumn{4}{c}{VSI-Bench} & \\
\cmidrule(lr){2-5}\cmidrule(lr){6-9}\cmidrule(lr){10-13}
Method & 4.5\% & 9\% & 14\% & 23\%
       & 4\% & 8\% & 17\% & 26\%
       & 4.5\% & 9\% & 14\% & 23\% & Avg. \\
\midrule
\bbmodel{14}{InternVL3-8B\ \ \normalfont\textcolor{black!55}{(3072 visual tokens, 12 views)}} \\
All tokens & \multicolumn{4}{c}{27.72} & \multicolumn{4}{c}{47.03} & \multicolumn{4}{c}{41.70} & 38.82 \\
\famllm{FastV} & -- & 18.07 & 21.05 & 25.26 & -- & 38.16 & 42.60 & 44.42 & -- & 34.15 & 36.15 & 38.96 & -- \\
\famllm{PyramidDrop} & -- & 17.93 & 19.64 & 24.02 & -- & 38.68 & 41.80 & 44.05 & -- & 33.15 & 36.89 & 38.46 & -- \\
\famdiv{DivPrune} & 22.35 & 24.24 & 25.69 & 26.48 & 43.34 & 44.27 & 45.47 & 46.21 & 37.40 & 38.69 & 39.95 & 40.41 & 36.21 \\
\famdiv{CDPruner} & 21.41 & 23.64 & 25.39 & 26.31 & 42.11 & 43.88 & 44.79 & 46.09 & 36.16 & 38.01 & 38.86 & 39.22 & 35.49 \\
\famimp{VisPruner} & 20.88 & 22.93 & 24.79 & 26.14 & 41.09 & 42.60 & 45.24 & 45.61 & 36.85 & 38.30 & 38.82 & 39.80 & 35.25 \\
\famgeo{SeGPruner} & 21.05 & 23.40 & 24.58 & 25.60 & 41.83 & 43.45 & 45.10 & 46.15 & 36.76 & 37.19 & 39.52 & 40.44 & 35.42 \\
\famgeo{CoVeR} & 22.31 & 25.01 & \underline{25.95} & \best{26.74} & 43.59 & 44.87 & 45.69 & 46.32 & \underline{38.53} & \underline{40.10} & \underline{41.05} & \best{42.14} & \underline{36.86} \\
\famgeo{3DZip} & \underline{23.59} & \underline{25.28} & 25.73 & 26.59 & \underline{43.65} & \underline{45.07} & \underline{45.75} & \best{46.60} & 38.22 & 39.40 & 39.99 & 41.19 & 36.76 \\
\rowcolor{tablecolor}\textbf{\ours{}} & \best{26.22} & \best{26.25} & \best{26.82} & \underline{26.63} & \best{45.64} & \best{45.44} & \best{46.46} & \underline{46.49} & \best{40.52} & \best{41.48} & \best{41.08} & \underline{41.69} & \best{37.89}\,{\scriptsize(\dgood{+1.03})} \\
\midrule
\bbmodel{14}{Qwen3-VL-8B\ \ \normalfont\textcolor{black!55}{(3600 visual tokens, 12 views)}} \\
All tokens & \multicolumn{4}{c}{27.32} & \multicolumn{4}{c}{44.19} & \multicolumn{4}{c}{54.25} & 41.92 \\
\famllm{FastV} & -- & 18.52 & 21.54 & 23.74 & -- & 35.44 & 40.01 & 41.66 & -- & 35.47 & 41.19 & 47.27 & -- \\
\famllm{PyramidDrop} & -- & 16.79 & 18.52 & 23.14 & -- & 34.64 & 39.07 & 41.69 & -- & 33.79 & 38.64 & 45.76 & -- \\
\famdiv{DivPrune} & \underline{23.29} & 24.88 & 25.22 & 25.52 & 40.86 & 41.57 & 42.51 & 43.05 & 44.04 & 47.55 & 48.78 & 50.46 & 38.14 \\
\famdiv{CDPruner} & 22.89 & 24.83 & 25.60 & 26.33 & 40.86 & \underline{42.28} & \underline{43.39} & 42.82 & 43.67 & 47.11 & 49.46 & 50.30 & 38.30 \\
\famimp{VisPruner} & 22.55 & 24.49 & \underline{26.03} & 26.31 & 40.10 & 41.38 & 42.94 & 42.82 & 41.44 & 44.55 & 47.53 & 50.61 & 37.56 \\
\famgeo{SeGPruner} & 22.67 & 24.68 & 25.28 & \underline{26.44} & 39.81 & 41.29 & 43.11 & \underline{43.82} & 41.45 & 46.72 & 49.00 & 51.10 & 37.95 \\
\famgeo{CoVeR} & 22.87 & \underline{25.45} & 25.58 & 26.27 & 40.86 & 42.03 & 43.25 & 43.34 & \underline{45.70} & \underline{48.96} & 50.21 & \best{52.90} & \underline{38.95} \\
\famgeo{3DZip} & 23.04 & 24.96 & 25.80 & 26.35 & \underline{41.57} & 42.17 & 43.14 & 43.45 & 45.69 & 48.63 & \underline{50.44} & 51.37 & 38.88 \\
\rowcolor{tablecolor}\textbf{\ours{}} & \best{26.12} & \best{26.33} & \best{26.72} & \best{26.91} & \best{43.25} & \best{44.36} & \best{44.53} & \best{44.50} & \best{49.56} & \best{51.38} & \best{51.89} & \underline{52.75} & \best{40.69}\,{\scriptsize(\dgood{+1.74})} \\
\midrule
\bbmodel{14}{Qwen3-VL-4B\ \ \normalfont\textcolor{black!55}{(3600 visual tokens, 12 views)}} \\
All tokens & \multicolumn{4}{c}{26.93} & \multicolumn{4}{c}{43.90} & \multicolumn{4}{c}{51.16} & 40.66 \\
\famllm{FastV} & -- & 16.15 & 18.65 & 21.41 & -- & 34.21 & 38.31 & 39.70 & -- & 32.51 & 36.13 & 42.70 & -- \\
\famllm{PyramidDrop} & -- & 15.34 & 16.86 & 20.68 & -- & 34.13 & 36.80 & 40.01 & -- & 31.05 & 34.45 & 41.34 & -- \\
\famdiv{DivPrune} & \underline{23.32} & 24.47 & 25.52 & 26.22 & 40.10 & \underline{42.26} & 42.51 & 42.77 & 42.69 & 46.38 & 47.44 & 48.85 & \underline{37.71} \\
\famdiv{CDPruner} & 23.19 & \underline{25.07} & 25.58 & 26.42 & 39.90 & 42.06 & 42.60 & \best{43.19} & 42.69 & 45.15 & 47.05 & 48.99 & 37.66 \\
\famimp{VisPruner} & 20.98 & 23.10 & 24.88 & 25.86 & 38.87 & 40.75 & 42.51 & 43.02 & 39.05 & 43.78 & 45.56 & 48.13 & 36.37 \\
\famgeo{SeGPruner} & 21.65 & 23.70 & 24.21 & 25.58 & 39.13 & 40.35 & \underline{42.65} & \underline{43.17} & 40.06 & 43.15 & 46.53 & 48.41 & 36.55 \\
\famgeo{CoVeR} & 21.73 & 24.71 & \underline{25.67} & \underline{26.44} & 39.56 & 41.18 & 42.11 & 42.91 & 42.81 & \underline{46.61} & 47.60 & \underline{49.67} & 37.58 \\
\famgeo{3DZip} & 22.82 & 24.11 & 24.98 & 26.14 & \underline{40.21} & 41.29 & 42.40 & \underline{43.17} & \underline{43.51} & 46.16 & \underline{47.87} & 49.18 & 37.65 \\
\rowcolor{tablecolor}\textbf{\ours{}} & \best{26.18} & \best{26.59} & \best{26.57} & \best{27.06} & \best{43.34} & \best{43.08} & \best{43.08} & 42.85 & \best{49.62} & \best{49.71} & \best{50.32} & \best{50.41} & \best{39.90}\,{\scriptsize(\dgood{+2.19})} \\
\midrule
\bbmodel{14}{LLaVA-OV-7B\ \ \normalfont\textcolor{black!55}{(8748 visual tokens, 12 views)}} \\
All tokens & \multicolumn{4}{c}{28.32} & \multicolumn{4}{c}{51.61} & \multicolumn{4}{c}{33.51} & 37.81 \\
\famllm{FastV} & -- & 18.50 & 21.73 & 24.92 & -- & 38.56 & 44.44 & 48.14 & -- & 25.74 & 27.81 & 30.13 & -- \\
\famllm{PyramidDrop} & -- & 17.86 & 20.41 & 25.13 & -- & 39.56 & 44.56 & 48.34 & -- & 23.35 & 28.49 & 31.95 & -- \\
\famdiv{DivPrune} & 18.63 & 23.81 & 25.22 & 26.80 & 41.26 & 44.64 & 46.23 & 48.82 & 29.99 & 31.59 & 33.31 & 34.31 & 33.72 \\
\famdiv{CDPruner} & \underline{23.38} & 24.56 & 26.07 & 27.49 & 45.50 & 48.22 & 49.56 & 50.10 & 28.80 & 30.82 & 31.93 & 33.75 & 35.02 \\
\famimp{VisPruner} & 20.90 & 23.40 & 24.80 & 26.90 & 44.87 & 45.70 & 49.00 & 49.50 & 28.69 & 30.65 & 31.98 & 33.58 & 34.16 \\
\famgeo{SeGPruner} & 22.95 & 24.50 & 26.40 & 27.70 & \underline{46.92} & 48.40 & 49.70 & 50.60 & 29.51 & 31.51 & 32.82 & 33.58 & 35.38 \\
\famgeo{CoVeR} & 22.91 & \underline{27.10} & \underline{27.90} & \underline{28.50} & 45.27 & \underline{48.60} & \underline{51.20} & \underline{51.70} & 32.44 & \underline{34.69} & \underline{34.84} & \underline{34.62} & \underline{36.65} \\
\famgeo{3DZip} & 22.42 & 25.75 & 27.23 & 28.02 & 41.57 & 45.98 & 47.26 & 48.68 & \underline{32.82} & 34.68 & 34.76 & 34.59 & 35.31 \\
\rowcolor{tablecolor}\textbf{\ours{}} & \best{25.58} & \best{27.81} & \best{28.06} & \best{28.60} & \best{47.57} & \best{50.75} & \best{51.69} & \best{51.77} & \best{33.29} & \best{35.16} & \best{34.88} & \best{34.68} & \best{37.49}\,{\scriptsize(\dgood{+0.84})} \\
\bottomrule
\end{tabular*}
\end{table*}

\subsection{Main Results}
\label{sec:results}

\begin{figure}[t]
  \centering
  \includegraphics[width=\columnwidth]{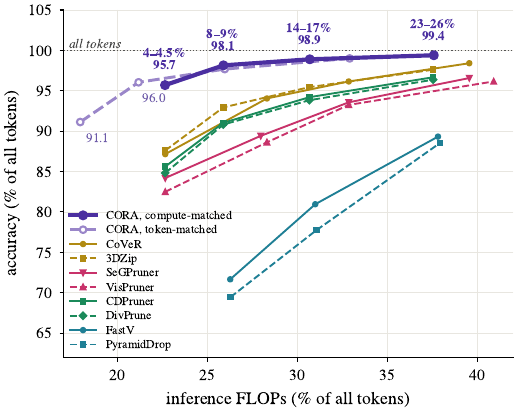}
  \caption{\textbf{Performance-efficiency trade-off curves}, normalized to full-token inference and averaged over four backbones and three benchmarks. FLOPs include the vision encoder and LLM prefill. \famlegend.}
  \label{fig:retained}
\end{figure}

\noindent\textbf{Accuracy across budgets.}
\ours{} achieves the highest score in 43 of 48 backbone--benchmark--budget settings, including all 12 settings at the tightest reference budgets (\cref{tab:adv_main}).
At these budgets, it retains 95.7\% of full-token performance on average, compared with 87.6\% for the strongest baseline, 3DZip (\cref{fig:retained}).
On VSI-Bench, its improvement over the strongest baseline ranges from 0.47 to 6.11 points across backbones, and on ScanQA it reaches 2.86 points on Qwen3-VL-4B.
Averaged over all three benchmarks and four budgets, \ours{} improves on the strongest baseline's mean score by 0.84--2.19 points across backbones, achieving the highest mean score on every backbone.
At the 8--9\% reference budgets, it retains 98.1\% of full-token performance, compared with 94.2\% for the strongest baseline, CoVeR.
Qualitative examples are shown in \cref{fig:qual_ivl}.

\noindent\textbf{Transfer to video.}
We test \ours{} beyond static RGB-D scans on VideoMME~\cite{Fu_2025_CVPR} and MVBench~\cite{Li_2024_CVPR}, using 12 uniformly sampled frames and VGGT-predicted geometry~\cite{Wang_2025_CVPR} for all methods.
We compare with the 3D-specific baselines, which also use geometry (\cref{tab:video}).
\ours{} achieves the highest score in 15 of 16 settings, including all 8 at the 4.5\% budget.
At 4.5\%, it retains 91.7\% of full-token accuracy, compared with at most 84.6\% for the baselines.
As in the 3D benchmarks, the gain over the strongest baseline grows at the smaller budget, from 2.40 points at 8\% to 3.82 points at 4.5\%.

\begin{table*}[t]
\centering
\tablefontsize
\setlength{\tabcolsep}{2pt}
\caption{\textbf{Video QA with VGGT-predicted geometry.} Matched encoder-plus-prefill FLOPs. Shaded: best; underlined: second best.}
\label{tab:video}
\begin{tabular*}{\textwidth}{@{\extracolsep{\fill}}l *{16}{c}@{}}
\toprule
 & \multicolumn{4}{c}{\textbf{InternVL3-8B}} & \multicolumn{4}{c}{\textbf{Qwen3-VL-8B}} & \multicolumn{4}{c}{\textbf{Qwen3-VL-4B}} & \multicolumn{4}{c}{\textbf{LLaVA-OV-7B}} \\
\cmidrule(lr){2-5}\cmidrule(lr){6-9}\cmidrule(lr){10-13}\cmidrule(lr){14-17}
 & \multicolumn{2}{c}{VideoMME} & \multicolumn{2}{c}{MVBench} & \multicolumn{2}{c}{VideoMME} & \multicolumn{2}{c}{MVBench} & \multicolumn{2}{c}{VideoMME} & \multicolumn{2}{c}{MVBench} & \multicolumn{2}{c}{VideoMME} & \multicolumn{2}{c}{MVBench} \\
\cmidrule(lr){2-3}\cmidrule(lr){4-5}\cmidrule(lr){6-7}\cmidrule(lr){8-9}\cmidrule(lr){10-11}\cmidrule(lr){12-13}\cmidrule(lr){14-15}\cmidrule(lr){16-17}
Method & 4.5\% & 8\% & 4.5\% & 8\% & 4.5\% & 8\% & 4.5\% & 8\% & 4.5\% & 8\% & 4.5\% & 8\% & 4.5\% & 8\% & 4.5\% & 8\% \\
\midrule
All tokens & \multicolumn{2}{c}{58.85} & \multicolumn{2}{c}{71.53} & \multicolumn{2}{c}{59.30} & \multicolumn{2}{c}{65.47} & \multicolumn{2}{c}{55.52} & \multicolumn{2}{c}{64.16} & \multicolumn{2}{c}{54.52} & \multicolumn{2}{c}{55.13} \\
\midrule
\famgeo{SeGPruner} & 50.33 & \underline{53.07} & 52.71 & 56.71 & 48.56 & \underline{50.81} & 51.45 & 55.37 & \underline{45.74} & \underline{49.96} & 51.39 & 54.08 & 49.11 & \underline{51.06} & 50.21 & 52.84 \\
\famgeo{CoVeR} & \underline{51.04} & 51.67 & \underline{56.97} & \underline{60.58} & \underline{48.85} & 50.52 & \underline{54.76} & \underline{56.55} & 44.81 & 46.48 & 54.47 & 56.37 & 46.81 & 48.48 & 51.18 & \best{53.79} \\
\famgeo{3DZip} & 49.93 & 52.67 & 56.68 & 60.16 & 48.07 & 49.44 & 52.89 & 55.39 & 44.48 & 46.85 & \underline{54.50} & \underline{56.47} & \underline{49.37} & 50.70 & \underline{52.32} & 53.13 \\
\rowcolor{tablecolor}\textbf{\ours{}} & \best{52.56} & \best{54.04} & \best{66.03} & \best{67.05} & \best{53.19} & \best{53.48} & \best{59.68} & \best{60.61} & \best{50.33} & \best{51.07} & \best{59.53} & \best{60.39} & \best{49.81} & \best{51.48} & \best{53.00} & \underline{53.39} \\
\bottomrule
\end{tabular*}
\end{table*}

\subsection{Analysis}
\label{sec:ablations}

\noindent\textbf{Token-matched and compute-matched allocation.}
\Cref{tab:paired_allocation} isolates the effect of reallocating encoder savings on the three backbones using Stage~1, across three benchmarks and the two tightest budgets. Both CORA configurations use the same selection rules; only their token allocations differ. The strongest baseline is selected separately for each setting from \cref{tab:adv_main}. With the token-matched allocation $(M,B)$, \ours{} outperforms this baseline in 14 of 18 settings while using 17.3--26.4\% fewer FLOPs. Reallocating the savings to $(M',B')$ improves all 18 settings by 1.76 points on average and outperforms the baseline throughout at approximately matched FLOPs (\cref{tab:paired_allocation_detail}). 

\begin{table}[!t]
\centering
\tablefontsize
\setlength{\tabcolsep}{3pt}
\caption{\textbf{Reallocating encoder savings.} Mean scores and encoder-plus-prefill TFLOPs over three benchmarks and the two tightest budgets. Baseline: strongest method in each setting.}
\label{tab:paired_allocation}
\begin{tabular*}{\columnwidth}{@{\extracolsep{\fill}}lrrrrrr@{}}
\toprule
& \multicolumn{2}{c}{Token-matched} & \multicolumn{2}{c}{Compute-matched} & \multicolumn{2}{c}{Baseline}\\
\cmidrule(lr){2-3}\cmidrule(lr){4-5}\cmidrule(l){6-7}
Backbone & Score & FLOPs & Score & FLOPs & Score & FLOPs\\
\midrule
InternVL3-8B & 36.79 & 10.22 & \best{37.59} & 12.59 & 36.04 & 12.60\\
Qwen3-VL-8B & 38.10 & 14.60 & \best{40.17} & 18.66 & 37.88 & 18.66\\
Qwen3-VL-4B & 37.34 & 9.46 & \best{39.75} & 12.61 & 36.83 & 12.61\\
\bottomrule
\end{tabular*}
\end{table}

\begin{table}[!t]
\centering
\tablefontsize
\setlength{\tabcolsep}{1pt}
\caption{\textbf{Token-matched ($M$) and compute-matched ($M'$) \ours{}.} Encoder-plus-prefill costs in TFLOPs. Shaded: best; underlined: second best.}
\label{tab:paired_allocation_detail}
\begin{tabular*}{\columnwidth}{@{\extracolsep{\fill}}llrrrrlrr@{}}
\toprule
& & \multicolumn{2}{c}{\ours{} ($M$)} & \multicolumn{2}{c}{\ours{} ($M'$)} & \multicolumn{3}{c}{Strongest baseline}\\
\cmidrule(lr){3-4}\cmidrule(lr){5-6}\cmidrule(l){7-9}
Dataset & Budget & Score & FLOPs & Score & FLOPs & Method & Score & FLOPs\\
\midrule
\bbmodel{9}{InternVL3-8B}\\
ScanQA & 4.5\% & \underline{24.28} & 9.44 & \best{26.22} & 11.78 & \famgeo{3DZip} & 23.59 & 11.79 \\
ScanQA & 9\% & \underline{25.88} & 11.25 & \best{26.25} & 13.60 & \famgeo{3DZip} & 25.28 & 13.60 \\
SQA3D & 4\% & \underline{44.16} & 9.20 & \best{45.64} & 11.59 & \famgeo{3DZip} & 43.65 & 11.59 \\
SQA3D & 8\% & \underline{45.41} & 10.82 & \best{45.44} & 13.20 & \famgeo{3DZip} & 45.07 & 13.21 \\
VSI-Bench & 4.5\% & \underline{39.59} & 9.40 & \best{40.52} & 11.78 & \famgeo{CoVeR} & 38.53 & 11.79 \\
VSI-Bench & 9\% & \underline{41.43} & 11.22 & \best{41.48} & 13.60 & \famgeo{CoVeR} & 40.10 & 13.60 \\
\midrule
\bbmodel{9}{Qwen3-VL-8B}\\
ScanQA & 4.5\% & \underline{24.04} & 13.64 & \best{26.12} & 17.64 & \famdiv{DivPrune} & 23.29 & 17.65 \\
ScanQA & 9\% & 25.41 & 15.93 & \best{26.33} & 19.93 & \famgeo{CoVeR} & \underline{25.45} & 19.93 \\
SQA3D & 4\% & 41.26 & 13.30 & \best{43.25} & 17.39 & \famgeo{3DZip} & \underline{41.57} & 17.40 \\
SQA3D & 8\% & \underline{42.65} & 15.33 & \best{44.36} & 19.42 & \famdiv{CDPruner} & 42.28 & 19.42 \\
VSI-Bench & 4.5\% & \underline{46.33} & 13.54 & \best{49.56} & 17.64 & \famgeo{CoVeR} & 45.70 & 17.65 \\
VSI-Bench & 9\% & 48.94 & 15.83 & \best{51.38} & 19.92 & \famgeo{CoVeR} & \underline{48.96} & 19.93 \\
\midrule
\bbmodel{9}{Qwen3-VL-4B}\\
ScanQA & 4.5\% & \underline{23.57} & 9.02 & \best{26.18} & 12.12 & \famdiv{DivPrune} & 23.32 & 12.13 \\
ScanQA & 9\% & \underline{25.13} & 10.11 & \best{26.59} & 13.22 & \famdiv{CDPruner} & 25.07 & 13.21 \\
SQA3D & 4\% & \underline{41.26} & 8.84 & \best{43.34} & 12.00 & \famgeo{3DZip} & 40.21 & 12.01 \\
SQA3D & 8\% & 42.09 & 9.80 & \best{43.08} & 12.97 & \famdiv{DivPrune} & \underline{42.26} & 12.97 \\
VSI-Bench & 4.5\% & \underline{44.50} & 8.95 & \best{49.62} & 12.13 & \famgeo{3DZip} & 43.51 & 12.13 \\
VSI-Bench & 9\% & \underline{47.52} & 10.03 & \best{49.71} & 13.22 & \famgeo{CoVeR} & 46.61 & 13.21 \\
\bottomrule
\end{tabular*}
\end{table}

\begin{table}[t]
\centering
\tablefontsize
\setlength{\tabcolsep}{1.5pt}
\renewcommand{\arraystretch}{1.15}
\caption{\textbf{Stage ablation at the lowest budget}, averaged over three benchmarks. Changes are relative to full \ours{}. \xmark: removed; \textit{rand}: random selection; Lower: configurations with reduced accuracy.}
\label{tab:cora_ablation_matched}
\begin{tabular*}{\columnwidth}{@{\extracolsep{\fill}}ccc rrrrr r@{}}
\toprule
\multicolumn{3}{c}{Stage} & \multicolumn{4}{c}{Backbone} & \multicolumn{2}{c}{Summary} \\
\cmidrule(lr){1-3}\cmidrule(lr){4-7}\cmidrule(l){8-9}
\textcolor{stage0}{1} & \textcolor{stageA}{2} & \textcolor{stageB}{3} & \shortstack{InternVL3\\8B} & \shortstack{Qwen3\\VL-8B} & \shortstack{Qwen3\\VL-4B} & \shortstack{LLaVA\\OV-7B} & Mean & Lower \\
\midrule
\rowcolor{groupband}\multicolumn{9}{@{}l}{\textit{Budget: ScanQA and VSI-Bench 4.5\%, SQA3D 4\%}} \\
\rowcolor{tablecolor}\cmark & \cmark & \cmark & \textbf{37.46} & \textbf{39.64} & \textbf{39.71} & \textbf{35.48} & \textbf{38.07} & \\
\xmark & \cmark & \cmark & \dbad{-1.37} & \dbad{-2.81} & \dbad{-4.41} & -- & \dbad{-2.86} & 9/9 \\
\cmark & \xmark & \cmark & \dbad{-0.70} & \dbad{-1.14} & \dbad{-2.39} & \dbad{-11.75} & \dbad{-3.99} & 12/12 \\
\cmark & \cmark & \xmark & \dbad{-0.05} & \dbad{-0.42} & \dbad{-0.84} & \dbad{-1.58} & \dbad{-0.72} & 11/12 \\
\cmark & \textit{rand} & \cmark & \dbad{-0.16} & \dneutral{+0.16} & \dbad{-0.42} & \dbad{-4.08} & \dbad{-1.12} & 8/12 \\
\cmark & \cmark & \textit{rand} & \dbad{-0.81} & \dbad{-0.58} & \dbad{-0.72} & \dbad{-1.05} & \dbad{-0.79} & 12/12 \\
\cmark & \xmark & \xmark & \dbad{-0.51} & \dbad{-0.28} & \dbad{-0.74} & \dbad{-13.95} & \dbad{-3.87} & 10/12 \\
\xmark & \cmark & \xmark & \dbad{-2.28} & \dbad{-4.28} & \dbad{-5.27} & \dbad{-1.58} & \dbad{-3.35} & 12/12 \\
\xmark & \xmark & \cmark & \dbad{-5.26} & \dbad{-9.55} & \dbad{-11.35} & \dbad{-11.75} & \dbad{-9.48} & 12/12 \\
\bottomrule

\end{tabular*}
\end{table}

\noindent\textbf{Stage contributions to accuracy and computation.}
We evaluate stage removals at the lowest reference budget (\cref{tab:cora_ablation_matched}).
Removing Stage~1 restores the full encoder and adjusts the language-model token counts to match total computation.
Removing Stage~2 or Stage~3 preserves the layer-averaged language-model token budget.
Random-selection controls retain the original token schedule and average three seeds.
We separately remove stages without adjusting the remaining schedule to measure their computational savings over the two tightest budgets.

Removing Stage 1 reduces the mean score by 2.86 points across the nine applicable backbone-benchmark settings. Under the matched language-model token budget, removing Stage 2 reduces the mean score by 3.99 points across 12 settings. Without compensatory pruning, removing it increases total FLOPs by 123.7\% for a mean score improvement of only 0.34 points. With Stage 1 enabled, replacing Stage 2 with random selection reduces the mean score by only 0.14 points, but on LLaVA-OneVision, which omits Stage 1, it reduces the score by 4.08 points. Removing Stages 1 and 2 together produces the largest mean decrease, 9.48 points.

Stage~3 reduces later-layer computation with little change in accuracy under the original allocation.
Removing it without compensatory pruning increases FLOPs by 11.3\% while changing the mean score by only $+0.02$ points.
Under the matched language-model token budget, however, removing Stage~3 reduces the mean score by 0.72 points.
Replacing its attention-based selection with random selection at the same token counts reduces the mean score by 0.79 points, with decreases in all 12 settings.
This comparison supports the contribution of question-conditioned selection beyond the allocation of tokens across layers.

\noindent\textbf{Scene selection under progressive allocation.}
We evaluate six baseline selectors under CORA's progressive token schedule with Stage~3 enabled at matched compute (\cref{tab:selector_schedule}).
\ours{} outperforms the strongest alternative in 22 of 24 settings, with larger gains at tighter budgets. For encoder-pruned backbones, this jointly evaluates scene selection and encoder-savings reallocation. On LLaVA-OneVision, where $M'=M$, it isolates scene selection, with \ours{} winning 5 of 6 settings.

\begin{table}[!t]
\centering
\tablefontsize
\setlength{\tabcolsep}{1.5pt}
\caption{\textbf{Scene selection with progressive allocation.} Score differences (points) from the strongest of six selectors with Stage~3 at matched compute; selection and allocation vary jointly.}
\label{tab:selector_schedule}
\begin{tabular*}{\columnwidth}{@{\extracolsep{\fill}}l rr rr rr@{}}
\toprule
 & \multicolumn{2}{c}{ScanQA} & \multicolumn{2}{c}{SQA3D} & \multicolumn{2}{c}{VSI-Bench} \\
\cmidrule(lr){2-3}\cmidrule(lr){4-5}\cmidrule(l){6-7}
 & 4.5\% & 9\% & 4\% & 8\% & 4.5\% & 9\% \\
\midrule
InternVL3-8B & \dgood{+2.74} & \dgood{+0.77} & \dgood{+1.76} & $-0.28$ & \dgood{+0.21} & \dgood{+1.01} \\
Qwen3-VL-8B & \dgood{+1.86} & \dgood{+0.43} & \dgood{+1.65} & \dgood{+1.48} & \dgood{+2.57} & \dgood{+1.14} \\
Qwen3-VL-4B & \dgood{+2.12} & \dgood{+1.18} & \dgood{+1.93} & \dgood{+0.45} & \dgood{+4.31} & \dgood{+1.39} \\
LLaVA-OV-7B & \dgood{+0.88} & \dgood{+0.64} & \dgood{+0.85} & \dgood{+2.02} & $-0.60$ & \dgood{+0.29} \\
\addlinespace[1pt]
Mean & \dgood{+1.90} & \dgood{+0.75} & \dgood{+1.55} & \dgood{+0.92} & \dgood{+1.62} & \dgood{+0.96} \\
\bottomrule
\end{tabular*}
\end{table}

\begin{table}[!t]
\centering
\tablefontsize
\setlength{\tabcolsep}{3pt}
\caption{\textbf{Robustness at the lowest reference budget.} Mean score changes (points) averaged over three benchmarks.}
\label{tab:geo_robust}
\begin{tabular*}{\columnwidth}{@{\extracolsep{\fill}}lrrrr@{}}
\toprule
Perturbation & \shortstack{InternVL3\\8B} & \shortstack{Qwen3\\VL-8B} & \shortstack{Qwen3\\VL-4B} & \shortstack{LLaVA\\OV-7B}\\
\midrule
Depth noise ($\sigma{=}0.2$) & \dneutral{+0.01} & \dbad{-0.15} & \dbad{-0.11} & \dbad{-0.79}\\
Missing depth (50\%) & \dbad{-0.04} & \dbad{-0.12} & \dbad{-0.42} & \dbad{-0.86}\\
Pose error (10\,cm, 5$^\circ$) & \dneutral{+0.17} & \dneutral{+0.02} & \dbad{-0.30} & \dbad{-0.46}\\
Pose error (25\,cm, 15$^\circ$) & \dneutral{+0.52} & \dneutral{+0.01} & \dbad{-0.24} & \dbad{-0.85}\\
Combined & \dbad{-0.20} & \dbad{-0.12} & \dbad{-0.74} & \dbad{-0.97}\\
VGGT (oracle scale) & \dbad{-0.06} & \dneutral{+0.07} & \dbad{-0.28} & \dbad{-1.44}\\
\rowcolor{tablecolor}VGGT (median scale) & \dbad{-0.23} & \dneutral{+0.23} & \dbad{-0.16} & \dbad{-1.11}\\
6 of 12 views, \ours{} & \dbad{-0.87} & \dbad{-2.04} & \dbad{-2.65} & \dbad{-0.30}\\
6 of 12 views, unpruned & \dbad{-1.94} & \dbad{-3.63} & \dbad{-3.19} & \dbad{-1.06}\\
\bottomrule
\end{tabular*}
\end{table}

\begin{table}[t]
\centering
\tablefontsize
\setlength{\tabcolsep}{2pt}
\caption{\textbf{End-to-end efficiency on ScanQA at the 9\% reference budget}, averaged over three models. Speedup is the mean of the per-backbone ratios to full-token inference.}
\label{tab:eff}\label{tab:cora_e2e_eff}
\resizebox{\columnwidth}{!}{%
\begin{tabular}{@{}lrrrrrrr@{}}
\toprule
Method & EM@1 & \shortstack{Pre-LLM\\(ms)} & \shortstack{Prefill\\(ms)} & \shortstack{TTFT\\(ms)} & \shortstack{Total\\(ms)} & \shortstack{Speedup\\($\times$)} & \shortstack{Peak\\(GiB)} \\
\midrule
All tokens & 27.79 & 218 & 889 & 1107 & 1175 & 1.00 & 19.3 \\
\midrule
\famllm{FastV} & 18.36 & 218 & 102 & 320 & 383 & 3.41 & 16.9 \\
\famllm{PyramidDrop} & 17.53 & 215 & 111 & 326 & 384 & 3.33 & 16.9 \\
\famdiv{DivPrune} & 24.31 & 199 & 107 & 306 & 386 & 3.54 & 16.8 \\
\famdiv{CDPruner} & 24.34 & 200 & 107 & 307 & 374 & 3.53 & 16.8 \\
\famimp{VisPruner} & 23.61 & 233 & 106 & 338 & 399 & 3.22 & 18.4 \\
\famgeo{SeGPruner} & 24.19 & 227 & 106 & 333 & 392 & 3.28 & 18.4 \\
\famgeo{CoVeR} & 25.85 & 243 & 105 & 348 & 412 & 3.11 & 16.8 \\
\famgeo{3DZip} & 25.33 & 216 & 108 & 324 & 396 & 3.35 & 16.8 \\
\rowcolor{tablecolor}\textbf{\ours{}} & \textbf{26.80} & 200 & \textbf{102} & \textbf{302} & \textbf{364} & \textbf{3.58} & \textbf{16.4} \\
\bottomrule
\end{tabular}}
\end{table}

\noindent\textbf{Robustness to imperfect geometry and missing views.}
We evaluate two aspects of robustness: sensitivity to the geometry used for token selection and sensitivity to reduced visual coverage (\cref{tab:geo_robust}).
First, we perturb depth and camera poses while holding images, prompts, and the reference budget $K$ fixed, and separately evaluate VGGT-predicted geometry.
Across these geometry conditions, the mean score change over four backbones and three benchmarks ranges from $-0.14$ to $-0.51$ points.
On the three backbones using Stage~1, the largest decrease is 1.39 points under synthetic perturbations and 0.88 points with predicted geometry.
LLaVA-OneVision exhibits greater sensitivity, with decreases of up to 2.10 points.
Second, we remove half the views and measure each model's score change relative to its own 12-view result.
CORA's accuracy decrease is no greater than that of the unpruned model in all 12 backbone--benchmark pairs.
These results indicate that CORA tolerates the evaluated geometry errors and does not amplify the accuracy degradation caused by the tested view reduction.
Complete results are provided in the supplementary material.

\noindent\textbf{Efficiency.}
\Cref{tab:eff} reports measured efficiency on ScanQA at the 9\% reference budget, averaged over three backbones. Under the scene-reuse protocol, initial selection costs are included and amortized across questions, while the encoder and LLM run for every question (details in the supplementary material). \ours{} achieves the highest accuracy, lowest mean time to first token (TTFT; 302\,ms), lowest total latency, and lowest peak memory (16.4\,GiB). Its mean per-backbone TTFT speedup over full-token inference is $3.58\times$.

\vspace{1mm}
\section{Conclusion}
\label{sec:conclusion}
We presented \ours{}, a training-free framework that prunes visual tokens at three locations. It removes repeated surface observations inside the vision encoder, selects a diverse scene context before the language model, and keeps question-relevant tokens in later layers. Across four backbones and three 3D benchmarks, \ours{} leads in 43 of 48 settings, retaining 95.7\% of full-token performance at tight budgets. Ablations and video transfer validate coordinated compression. Future work includes geometry-free selection, support for single-patch tokens, improved robustness, and faster scene selection to reduce overhead.

\section*{LLM Usage Statement}
Claude (Anthropic) and Codex (OpenAI) were used during manuscript preparation. This assistance included language editing and rephrasing, LaTeX formatting, figure and table organization, and consistency checks of numerical summaries against existing result tables. The authors take full responsibility for the entire content of the paper, including all scientific claims, reported results, and references.

{
    \small
    \bibliographystyle{legacy/ieeenat_fullname}
    \bibliography{main}
}
\clearpage
\thispagestyle{empty}
\maketitlesupplementary
\appendix
\setcounter{figure}{0}
\setcounter{table}{0}
\setcounter{equation}{0}
\renewcommand{\thefigure}{A\arabic{figure}}
\renewcommand{\thetable}{A\arabic{table}}
\renewcommand{\theequation}{A\arabic{equation}}

\noindent
This supplementary material reports the implementation settings, the compute calibration, and the complete results that the main paper summarizes: the unmatched-budget and per-benchmark ablations behind the stage analysis (\cref{sec:supp_ablation}), the per-benchmark geometry-robustness grid (\cref{sec:supp_robust}), and the timing protocol (\cref{sec:supp_timing}). 

\section{Implementation details}
\label{sec:supp_impl}

\noindent\textbf{Hyperparameters.}
\Cref{tab:supp_hparams} lists the hyperparameters of \ours{}: the six quantities that we choose and that change which tokens are selected. We use the same values for all backbones, budgets, and benchmarks, and tune none of them per setting.

\noindent\textbf{Allocation rule.}
The token schedule is fixed by the rule of \cref{sec:supp_budget} rather than tuned: the late-to-early ratio is $r=0.5$, question-conditioned selection follows layer $L=\lfloor16n_L/28\rfloor$, and Stage~1 prunes after encoder layer $\ell_0=\mathrm{round}(d/2)$. 

\noindent\textbf{Implementation constants.}
Four further constants are properties of the data or of the solver rather than tuned quantities. Depths outside $(0.2,6)$\,m are treated as invalid, the sensor-validity window that every evaluated method uses. The neighborhood radius has a floor of 0.02\,m, which guards against a degenerate radius when the coverage seeds coincide.
Each candidate is assigned at most 256 nearest neighbors, and each round of the greedy fill adds at most 16 candidates; both bound the cost of the selection. 

\begin{table}[h]
\centering
\tablefontsize
\setlength{\tabcolsep}{4pt}
\caption{\textbf{Hyperparameters of \ours{}.} The same values are used for all backbones, budgets, and benchmarks. The allocation rule and the implementation constants are given in the text.}
\label{tab:supp_hparams}
\begin{tabular*}{\columnwidth}{@{\extracolsep{\fill}}lc@{}}
\toprule
Hyperparameter & Value \\
\midrule
\rowcolor{groupband}\multicolumn{2}{@{}l}{\stagebadge{stage0}{1}\ \textcolor{stage0}{\textbf{Geometric coverage}}} \\
coverage target $\eta$ & 0.95 \\
voxel size & 0.05\,m \\
\rowcolor{groupband}\multicolumn{2}{@{}l}{\stagebadge{stageA}{2}\ \textcolor{stageA}{\textbf{Representative scene selection}}} \\
seed fraction $\alpha$ & 0.4 \\
spatial weight $\gamma$ & 0.5 \\
neighborhood multiplier ($R=2d_s$) & 2 \\
multi-view radius for $w_i$ & 0.1\,m \\
\bottomrule
\end{tabular*}
\end{table}

\noindent\textbf{Per-backbone settings.}
\Cref{tab:supp_backbone} lists the encoder split and the Stage~3 layer of each backbone.
The encoder runs the layers up to the split on all units and the remaining layers on the retained set $\mathcal{U}$.
LLaVA-OneVision maps one $14$-pixel patch to each language-model token and therefore omits Stage~1, as stated in the main paper.

\begin{table}[h]
\centering
\tablefontsize
\setlength{\tabcolsep}{3pt}
\caption{\textbf{Per-backbone settings.} Encoder split: $\ell_0$ of $d$ encoder layers.}
\label{tab:supp_backbone}
\begin{tabular*}{\columnwidth}{@{\extracolsep{\fill}}lcccc@{}}
\toprule
Backbone & Encoder split & LLM layers $n_L$ & $L$ & Stage~\textcolor{stage0}{1} \\
\midrule
InternVL3-8B \cite{zhu2025internvl3exploringadvancedtraining} & 12 / 24 & 28 & 16 & \cmark \\
Qwen3-VL-8B \cite{bai2025qwen3vltechnicalreport} & 14 / 27 & 36 & 20 & \cmark \\
Qwen3-VL-4B \cite{bai2025qwen3vltechnicalreport} & 12 / 24 & 36 & 20 & \cmark \\
LLaVA-OV-7B \cite{li2024llavaonevisioneasyvisualtask} & -- & 28 & 16 & \xmark \\
\bottomrule
\end{tabular*}
\end{table}

\section{Evaluation metrics}
\label{sec:supp_metrics}
\noindent\textbf{ScanQA \cite{Azuma_2022_CVPR} and SQA3D \cite{ma2023sqa3dsituatedquestionanswering}.} We report EM@1: the percentage of questions whose top-1 answer exactly matches one of the reference answers.

\noindent\textbf{VSI-Bench \cite{yang2025thinkingspacemultimodallarge}.} We use the official metric code of VSI-Bench and report its overall score on the ScanNet and ARKitScenes subsets (3{,}672 questions). The benchmark has eight task types.
Multiple-choice tasks (relative direction, relative distance, route planning, appearance order) are scored by accuracy.
Numerical tasks (object counting, absolute distance, object size, room size) are scored by mean relative accuracy (MRA),
\begin{equation}
\mathrm{MRA}=\frac{1}{|\mathcal{C}_\theta|}\sum_{\theta\in\mathcal{C}_\theta}\mathbf{1}\!\left[\frac{|\hat y-y|}{y}<1-\theta\right],
\label{eq:supp_mra}
\end{equation}
with $\mathcal{C}_\theta=\{0.50,0.55,\dots,0.95\}$, where $\hat y$ is the predicted and $y$ the ground-truth value; an answer that cannot be parsed as a number scores 0.
Relative direction is the mean of its easy, medium, and hard splits.
The overall score is the unweighted mean of the eight task scores.

\section{Budgets and compute calibration}
\label{sec:supp_budget}

\noindent\textbf{Token-matched reference.}
With $M$ tokens in layers $1,\dots,L$ and $B=\mathrm{round}(rM)$ in the later layers, the exact layer average is
$\bar n(M,B)=(LM+(n_L-L)B)/n_L$.
Setting its unrounded approximation to the reference count $K$ gives
\begin{equation}
 M=\mathrm{round}\!\left(\frac{n_LK}{L+(n_L-L)r}\right),
 \quad B=\mathrm{round}(rM),
 \label{eq:supp_budget}
\end{equation}
with rounding half to even.
For the 28-layer models, $L=16$ and $M=\mathrm{round}(14K/11)$; for the 36-layer models, $L=20$ and $M=\mathrm{round}(9K/7)$.
This allocation is the token-matched reference; the main paper's results use the compute-matched counts $(M',B')$.

\noindent\textbf{Compute model.}
The calibration of \cref{eq:adv_compute_budget} uses estimated vision-encoder and LLM-prefill FLOPs, excluding decoding, token-selection operations, and the projector.
For $V=12$ views with $n$ encoder patches each, encoder depth $d$, hidden width $h_v$, and feed-forward width $h_{\rm ff,v}$, the full-encoder estimate is
\begin{equation}
 E_{\rm enc}=dV\bigl[2n(4h_v^2+2h_vh_{\rm ff,v})+4n^2h_v\bigr],
 \label{eq:supp_encoder_flops}
\end{equation}
which counts the encoder's attention term $4n^2h_v$.
The retained fraction $f$ scales this estimate using the measured per-layer token counts.
For LLM hidden width $h$, $n_{\rm kv}$ key/value heads, head dimension $h_a$, and feed-forward width $h_{\rm ff}$, the calibration uses
\begin{equation}
\begin{aligned}
 F_{\rm LLM}(m,b;L)&=L\,g(m)+(n_L-L)\,g(b),\\
 g(x)&=(x+100)c_{\rm lin}+2(x+100)^2h,\\
 c_{\rm lin}&=2(2h^2+2h n_{\rm kv}h_a+3h h_{\rm ff}).
\end{aligned}
\label{eq:supp_llm_flops}
\end{equation}
The 100 text tokens are a fixed assumption of this FLOPs model rather than the measured prompt length of every question.
The reference cost is a full encoder with the token-matched progressive schedule $(M,B,L)$.
Baseline costs are evaluated with their own token schedules.

\noindent\textbf{Calibration and granularity.}
For the static 3D benchmarks, $M'$ is calibrated per backbone, benchmark, and budget.
The encoder fraction is the mean per-question fraction measured in the $(\ell_0{=}d/2, M)$ runs at the two tightest budgets; the looser-budget calibration reuses the corresponding backbone--benchmark fraction.
The solve selects the largest integer $M'$ whose modeled cost does not exceed the reference. At the looser budgets, the Stage 1 floor $M'$ can bind and raise the encoder fraction above this estimate. We therefore recalibrated $M'$ per question from the geometric coverage count in cells whose cost exceeded the reference by more than 0.2\%.
For video, $M'$ is solved per question from the geometric coverage count, with the encoder fraction obtained from a fit to the corresponding backbone's $\ell_0{=}d/2$ image runs.
LLaVA-OneVision omits Stage~1 and therefore uses $M'=M$ and $B'=B$.

\noindent\textbf{Worked allocation.}
On InternVL3-8B ScanQA at the 9\% reference budget, $K=276$, $d=24$, $\ell_0=12$, $L=16$, and $n_L=28$.
The token-matched counts $M=351$ and $B=176$ give $\bar n=276.0$ and 11.25 TFLOPs.
The compute-matched counts $M'=577$ and $B'=288$ give $\bar n=453.1$ and 13.60 TFLOPs, against 13.61 TFLOPs for the full-encoder progressive reference.
The compute-matched allocation therefore spends the encoder savings on additional scene context instead of enforcing $\bar n=K$.
On LLaVA-OneVision ScanQA at 9\%, $K=787$ and $M'=M=1002$, $B'=B=501$, because Stage~1 is absent.

\noindent\textbf{Cost.}
The peak LLM visual-token count is $M'$, or the number of candidates when Stage~2 is candidate-limited.
With Stage~1 after $\ell_0=\mathrm{round}(d/2)$, the unselected units are excluded from encoder layers $\ell_0+1,\dots,d$.
Across the 36 compute-matched configurations on the three Stage~1 backbones, the encoder uses 69.4--74.9\% of the full-encoder FLOPs, a reduction of approximately 25--31\%.
The question-independent selection runs on the CPU once per scene; measured timing, including selection cost, is reported in \cref{tab:eff} and \cref{sec:supp_timing}.

\section{Stage 1 details}
\label{sec:supp_stage1}
\noindent\textbf{Candidate floor.}
If the greedy cover of \cref{eq:adv_cover} keeps fewer than $M'$ units, we first continue the greedy order until every observed voxel is covered, and then add the remaining units in index order until $M'$ units are kept.
Ties in the greedy order are broken by the lower index.
Formally, the floor is $m_{\min}=\min(T-|\mathcal{P}_{\rm enc}|,M')$.

\noindent\textbf{Padding.}
On InternVL3, the letterbox padding added when a $640\times480$ view is padded to a square has no depth, so it covers no voxel.
The encoder attends strongly to these rows, which act as attention sinks.
We therefore keep the outermost padding row at the top and at the bottom of each view (384 of the 768 padding tokens) in the late encoder layers, so that this sink remains available to the retained units, and drop the other padding tokens.
The kept padding tokens are removed after the encoder and are never candidates for Stage~2.
Qwen3-VL has no letterbox padding; LLaVA-OneVision does not use Stage~1, and its padding tokens are ordinary Stage~2 candidates.

\section{Stage 2 details}
\label{sec:supp_stage2}

\noindent\textbf{Token positions.}
Let $\Omega_t$ be the $P\times P$ pixel patch of token $t$ and $v(t)$ its view.
A pixel $\mathbf{p}=(u,w)$ of view $v$ with depth $z=D_v(\mathbf{p})$ back-projects to the world point
\begin{equation}
  \pi_v(\mathbf{p}, z) = E_v \Bigl[\tfrac{u-c_x}{f_x}\,z,\ \tfrac{w-c_y}{f_y}\,z,\ z,\ 1\Bigr]^{\!\top},
  \label{eq:supp_backproj}
\end{equation}
where the homogeneous coordinate is dropped after the multiplication. The intrinsics are those of the scene, rescaled to the model input with the letterbox padding taken into account; no model-specific constant is used. The mean back-projection of the patch of token $t$ is
\begin{equation}
  \bar{\mathbf{x}}_t = \frac{1}{P^2}\sum_{\mathbf{p}\in\Omega_t} \pi_{v(t)}\bigl(\mathbf{p}, D_{v(t)}(\mathbf{p})\bigr),
  \label{eq:supp_mean}
\end{equation}
in which pixels without depth ($D_v(\mathbf{p})=0$) contribute the camera center. Let
$\rho_t = |\{\mathbf{p}\in\Omega_t : D_{v(t)}(\mathbf{p})>0\}|/P^2$ be the valid-depth ratio of token $t$,
$\mathbf{c}_t$ the center of $\Omega_t$, and $\tilde z_v$ the median of the valid depths of view $v$. The position of token $t$ is
\begin{equation}
  \mathbf{x}_t =
  \begin{cases}
    \pi_{v(t)}\bigl(\mathbf{c}_t, \tilde z_{v(t)}\bigr), & \rho_t < 0.5 \text{ and } t\notin\mathcal{P},\\[2pt]
    \bar{\mathbf{x}}_t, & \text{otherwise},
  \end{cases}
  \label{eq:supp_lift}
\end{equation}
where $\mathcal{P}$ is the set of letterbox-padding tokens.
Without the first case, tokens on windows and on dark or reflective surfaces would collapse onto the camera center.
Padding tokens take the camera center of their view as position.
A token is a Stage~2 candidate if $\mathbf{x}_t$ is finite and non-zero and $t\in\mathcal{U}\setminus\mathcal{P}_{\rm enc}$, so padding tokens are candidates only when Stage~1 is disabled.
If the number of candidates does not exceed $M'$, all candidates are kept.

\noindent\textbf{Seeds.}
Farthest-point sampling draws $\max(1,\mathrm{round}(\alpha M'))$ seeds and starts from the candidate nearest to the centroid of all candidates.
The seed spacing $d_s$ is the distance from the last seed to its nearest earlier seed; for a single seed it is the maximum candidate-to-seed distance, $d_s=\max_{t\in\mathcal{C}}\lVert\mathbf{x}_t-\mathbf{x}_{s_1}\rVert$.
The neighborhood radius is $R=\max(2d_s,0.02\,\mathrm{m})$.

\noindent\textbf{Weights.}
$w_i=1/m_i$, where $m_i$ is one plus the number of other views that contain a candidate within 0.1\,m of $\mathbf{x}_i$.

\noindent\textbf{Batched greedy.}
The marginal gain of candidate $j$ is $g_j=F(S)-F(S\cup\{j\})$, computed from the current cost $\min(\bar c,\min_{k\in S}c_{ik})$ of every candidate $i$ in \cref{eq:adv_fl}.
Each round adds up to 16 candidates in decreasing order of $g_j$, skips any candidate within $R/2$ of one already added in that round, and the last round adds only the remaining number.

\section{Stage 3 details}
\label{sec:supp_stage3}
The question set $\mathcal{Q}$ includes the answer options of multiple-choice questions and excludes template and formatting tokens.
The attention $A$ is the causal softmax over all keys, computed with the query and key projections of layer $L{+}1$ on the hidden states after layer $L$, and then restricted to the visual tokens.
If all scores are equal, the min-max normalized score is zero.
Stage~3 keeps $\min(B',|S|)$ tokens, and ties in the top-$B'$ selection are broken by the lower token index.
Removed tokens are dropped from the hidden states and from the key-value cache of every layer, so decoding attends only to $S'$.

\section{Additional ablation results}
\label{sec:supp_ablation}

\noindent\textbf{Removing a stage without compensating.}
The stage ablation of the main paper keeps the budget matched when a stage is removed. It matches total computation for Stage 1 and the layer-averaged token budget for Stages 2 and 3.
\Cref{tab:supp_unmatched} instead removes a stage and leaves the remaining schedule unchanged, which raises the cost and isolates the computation that the stage saves.
Removing either stage raises the cost with a small change in the mean score, so both stages save computation at similar accuracy.

\begin{table}[!t]
\centering
\tablefontsize
\setlength{\tabcolsep}{3pt}
\caption{\textbf{Removing a stage without a matched budget}, relative to full \ours{}; mean over three benchmarks and the two tightest budgets (24 settings). FLOPs: encoder plus LLM prefill.}
\label{tab:supp_unmatched}
\begin{tabular*}{\columnwidth}{@{\extracolsep{\fill}}lrrrr@{}}
\toprule
 & \multicolumn{2}{c}{Without Stage~\textcolor{stageA}{2}} & \multicolumn{2}{c}{Without Stage~\textcolor{stageB}{3}}\\\cmidrule(lr){2-3}\cmidrule(l){4-5}
Backbone & Score $\Delta$ & FLOPs $\Delta$ & Score $\Delta$ & FLOPs $\Delta$\\
\midrule
InternVL3-8B & \dneutral{+0.34} & \dbad{+38.0\%} & \dneutral{+0.27} & \dbad{+10.8\%}\\
Qwen3-VL-8B & \dneutral{+1.24} & \dbad{+37.7\%} & \dneutral{-0.05} & \dbad{+11.4\%}\\
Qwen3-VL-4B & \dneutral{+0.09} & \dbad{+19.8\%} & \dneutral{-0.03} & \dbad{+10.9\%}\\
LLaVA-OV-7B & \dneutral{-0.30} & \dbad{+399.3\%} & \dneutral{-0.13} & \dbad{+12.0\%}\\
\midrule
\rowcolor{groupband}Mean & \dneutral{+0.34} & \dbad{+123.7\%} & \dneutral{+0.02} & \dbad{+11.3\%}\\
\bottomrule
\end{tabular*}
\end{table}

\noindent\textbf{Per-benchmark stage ablation on LLaVA-OneVision.}
\ours{} applies only Stages~2--3 on LLaVA-OneVision, whose tokens each cover a single $14$-pixel patch.
\Cref{tab:supp_llava_s1} gives the per-benchmark scores behind the LLaVA-OneVision column of the main paper's stage ablation at matched compute and the tightest budget.
Removing Stage~3 costs 1.58 points, whereas the configurations that drop Stage~2 lose 11.75 and 13.95 points, which confirms that Stage~2 carries the multi-view redundancy removal on this backbone.

\begin{table}[!ht]
\centering
\tablefontsize
\setlength{\tabcolsep}{3pt}
\caption{\textbf{Per-benchmark stage ablation on LLaVA-OneVision-7B}, compute-matched at the tightest budget (ScanQA and VSI-Bench 4.5\%, SQA3D 4\%). \xmark: removed; Stage~1 is not applied on this backbone. The mean and $\Delta$ columns are the LLaVA-OV column of the main paper's stage ablation.}
\label{tab:supp_llava_s1}
\begin{tabular*}{\columnwidth}{@{\extracolsep{\fill}}cc rrrrr@{}}
\toprule
\multicolumn{2}{c}{Stage} & \multicolumn{3}{c}{Benchmark} & \multicolumn{2}{c}{Summary}\\
\cmidrule(lr){1-2}\cmidrule(lr){3-5}\cmidrule(l){6-7}
\textcolor{stageA}{2} & \textcolor{stageB}{3} & ScanQA & SQA3D & VSI & Mean & $\Delta$\\
\midrule
\rowcolor{tablecolor}\cmark & \cmark & \textbf{25.58} & \textbf{47.57} & \textbf{33.29} & \textbf{35.48} & --\\
\cmark & \xmark & 23.74 & 45.58 & 32.39 & 33.90 & \dbad{-1.58}\\
\xmark & \cmark & 10.07 & 34.64 & 26.48 & 23.73 & \dbad{-11.75}\\
\xmark & \xmark & 5.11 & 31.63 & 27.85 & 21.53 & \dbad{-13.95}\\
\bottomrule
\end{tabular*}
\end{table}

\begin{table}[!t]
\centering
\tablefontsize
\setlength{\tabcolsep}{2pt}
\caption{\textbf{Stage~1 pruning layer $\ell_0$} at the token-matched allocation $(M,B)$, with the same $M$ and $B$ in every column. Encoder FLOPs left: fraction of the full-encoder cost. The last column prunes after the encoder and therefore saves no encoder computation. Shaded column: the setting used by \ours{}.}
\label{tab:supp_layer}
\begin{tabular*}{\columnwidth}{@{\extracolsep{\fill}}llrr>{\columncolor{tablecolor}}rrr@{}}
\toprule
 & & \multicolumn{5}{c}{Stage~\textcolor{stage0}{1} pruning layer $\ell_0$} \\
\cmidrule(l){3-7}
Benchmark & Budget & $0$ & $d/4$ & \textbf{$\boldsymbol{d/2}$} & $3d/4$ & $d$ \\
\midrule
\bbmodel{7}{InternVL3-8B}\\
\rowcolor{groupband}\multicolumn{2}{@{}l}{\textit{Encoder FLOPs left}} & 45\% & 59\% & 73\% & 86\% & 100\% \\
ScanQA & 4.5\% & 22.93 & 23.23 & \textbf{24.28} & 23.96 & 24.11 \\
ScanQA & 9\% & 24.41 & 24.94 & \textbf{25.88} & 25.56 & 25.58 \\
SQA3D & 4\% & 43.62 & 43.56 & \textbf{44.16} & 43.82 & 44.39 \\
SQA3D & 8\% & 44.33 & 44.87 & \textbf{45.41} & 45.52 & 45.38 \\
VSI-Bench & 4.5\% & 39.09 & 39.73 & \textbf{39.59} & 39.50 & 40.21 \\
VSI-Bench & 9\% & 40.70 & 40.63 & \textbf{41.43} & 40.72 & 40.64 \\
\midrule
\bbmodel{7}{Qwen3-VL-8B}\\
\rowcolor{groupband}\multicolumn{2}{@{}l}{\textit{Encoder FLOPs left}} & 40\% & 55\% & 71\% & 84\% & 100\% \\
ScanQA & 4.5\% & 24.04 & 23.68 & \textbf{24.04} & 24.04 & 23.42 \\
ScanQA & 9\% & 24.75 & 24.88 & \textbf{25.41} & 26.14 & 25.88 \\
SQA3D & 4\% & 41.18 & 40.92 & \textbf{41.26} & 40.81 & 41.77 \\
SQA3D & 8\% & 42.17 & 42.43 & \textbf{42.65} & 43.05 & 43.19 \\
VSI-Bench & 4.5\% & 45.36 & 46.06 & \textbf{46.33} & 47.22 & 46.31 \\
VSI-Bench & 9\% & 48.43 & 48.65 & \textbf{48.94} & 49.10 & 49.32 \\
\bottomrule
\end{tabular*}
\end{table}

\begin{table*}[!h]
\centering
\tablefontsize
\setlength{\tabcolsep}{2.4pt}
\caption{\textbf{Robustness to geometry errors at the lowest reference budget} (ScanQA and VSI-Bench 4.5\%, SQA3D 4\%). Changes are relative to unperturbed \ours{} at the same $K$, with one noise draw. In the view-reduction block, each model is compared with its own 12-view score. $^{\ast}$: the 95\% confidence interval excludes zero. Intervals are computed for the \ours{} rows only. Mean: over all 12 backbone--benchmark cells of the row.}
\label{tab:supp_geo_robust_full}
\begin{tabular}{@{}lrrr|rrr|rrr|rrr|>{\columncolor{groupband}}r@{}}
\toprule
 & \multicolumn{3}{c|}{\textbf{InternVL3-8B}} & \multicolumn{3}{c|}{\textbf{Qwen3-VL-8B}} & \multicolumn{3}{c|}{\textbf{Qwen3-VL-4B}} & \multicolumn{3}{c|}{\textbf{LLaVA-OV-7B}} & \\
\cmidrule(lr){2-4}\cmidrule(lr){5-7}\cmidrule(lr){8-10}\cmidrule(lr){11-13}
Perturbation & ScanQA & SQA3D & VSI & ScanQA & SQA3D & VSI & ScanQA & SQA3D & VSI & ScanQA & SQA3D & VSI & \textbf{Mean} \\
\midrule
\rowcolor{tablecolor}Unperturbed \ours{} & 26.22 & 45.64 & 40.52 & 26.12 & 43.25 & 49.56 & 26.18 & 43.34 & 49.62 & 25.58 & 47.57 & 33.29 & \\
\rowcolor{groupband}\multicolumn{14}{@{}l}{\textit{Geometry perturbations}} \\
Depth noise ($\sigma{=}0.2$) & \dbad{-0.24} & \dbad{-0.09} & \dgood{+0.35} & \dbad{-0.39} & \dbad{-0.48} & \dgood{+0.41} & \dgood{+0.56} & \dbad{-0.71} & \dbad{-0.17} & \dbad{-0.34} & \dbad{-1.65^{\ast}} & \dbad{-0.37} & \dbad{-0.26} \\
Missing depth (50\%) & \dbad{-0.19} & \dbad{-0.06} & \dgood{+0.14} & \dgood{+0.24} & \dbad{-0.74} & \dgood{+0.15} & \dgood{+0.24} & \dbad{-0.20} & \dbad{-1.31^{\ast}} & \dbad{-1.11} & \dbad{-0.85} & \dbad{-0.61} & \dbad{-0.36} \\
Pose error (10\,cm, 5$^\circ$) & \dbad{-0.30} & \dgood{+0.23} & \dgood{+0.59} & \dbad{-0.17} & \dbad{-0.40} & \dgood{+0.62} & \dgood{+0.04} & \dbad{-0.23} & \dbad{-0.72} & \dbad{-0.51} & \dbad{-0.03} & \dbad{-0.85} & \dbad{-0.14} \\
Pose error (25\,cm, 15$^\circ$) & \dgood{+0.19} & \dgood{+0.28} & \dgood{+1.08^{\ast}} & \dbad{-0.04} & \dbad{-0.37} & \dgood{+0.45} & \dgood{+0.13} & \dbad{-0.11} & \dbad{-0.74} & \dbad{-0.90} & \dbad{-1.22} & \dbad{-0.44} & \dbad{-0.14} \\
Combined & \dbad{-0.19} & \dgood{+0.03} & \dbad{-0.43} & \dbad{-0.19} & \dbad{-1.02} & \dgood{+0.85} & \dgood{+0.13} & \dbad{-0.97^{\ast}} & \dbad{-1.39^{\ast}} & \dbad{-0.81} & \dbad{-1.71} & \dbad{-0.38} & \dbad{-0.51} \\
VGGT (oracle scale) & \dbad{-0.49} & \dgood{+0.06} & \dgood{+0.25} & \dbad{-0.30} & \dbad{-0.03} & \dgood{+0.54} & \dgood{+0.19} & \dbad{-0.54} & \dbad{-0.50} & \dbad{-1.16^{\ast}} & \dbad{-2.10^{\ast}} & \dbad{-1.05^{\ast}} & \dbad{-0.43} \\
VGGT (median scale) & \dbad{-0.68} & \dbad{-0.03} & \dgood{+0.01} & \dzero{0.00} & \dbad{-0.20} & \dgood{+0.90} & \dgood{+0.68} & \dbad{-0.88} & \dbad{-0.29} & \dbad{-1.16^{\ast}} & \dbad{-2.05^{\ast}} & \dbad{-0.12} & \dbad{-0.32} \\
\rowcolor{groupband}\multicolumn{14}{@{}l}{\textit{View reduction}} \\
6 of 12 views, \ours{} & \dbad{-1.30^{\ast}} & \dbad{-0.54} & \dbad{-0.76} & \dbad{-1.03} & \dbad{-1.65^{\ast}} & \dbad{-3.43^{\ast}} & \dbad{-1.18^{\ast}} & \dbad{-2.27^{\ast}} & \dbad{-4.50^{\ast}} & \dgood{+0.68} & \dbad{-1.02} & \dbad{-0.56} & \dbad{-1.46} \\
6 of 12 views, unpruned & \dbad{-1.39} & \dbad{-2.22} & \dbad{-2.21} & \dbad{-2.03} & \dbad{-2.50} & \dbad{-6.36} & \dbad{-1.67} & \dbad{-2.27} & \dbad{-5.64} & \dbad{-0.96} & \dbad{-1.17} & \dbad{-1.05} & \dbad{-2.46} \\
\bottomrule
\end{tabular}
\end{table*}

\noindent\textbf{Encoder pruning layer.}
\Cref{tab:supp_layer} varies the layer $\ell_0$ at which Stage~1 drops the redundant units, holding the token-matched allocation $(M,B)$ fixed so that only the encoder cost changes.
Relative to pruning at $3d/4$, pruning at $d/2$ gives the same mean score over the 12 settings while using about 13 percentage points less encoder computation.
Pruning at the encoder input ($\ell_0=0$) costs $0.70$ points on average, and pruning only after the full encoder, which saves no encoder computation, gains $0.07$ points.
Letting the retained units accumulate image context through half of the encoder therefore recovers the accuracy of late pruning at a substantially lower encoder cost, and we use $\ell_0=\mathrm{round}(d/2)$ throughout.
The $d/2$ column reproduces the token-matched \ours{} scores of \cref{tab:paired_allocation_detail}.

\section{Robustness: per-benchmark results}
\label{sec:supp_robust}
\Cref{tab:supp_geo_robust_full} gives the per-benchmark results behind the robustness analysis of the main paper, whose table reports their three-benchmark means.
Depth and camera poses are perturbed while the images, prompts, and reference budget $K$ are held fixed, with one noise draw per condition; the view-reduction block compares each model with its own 12-view score.
The rightmost column is the mean over the four backbones and three benchmarks, which ranges from $-0.14$ to $-0.51$ points across the geometry conditions.
On the three backbones using Stage~1, no synthetic perturbation costs more than 1.39 points and predicted geometry costs at most 0.88 points, whereas LLaVA-OneVision, which uses geometry only in Stage~2, loses up to 2.10 points.
With half of the views removed, \ours{} loses no more accuracy than the unpruned model in all 12 backbone--benchmark pairs.

\section{Timing and memory}
\label{sec:supp_timing}
We measure inference efficiency with a shared profiling harness on all 4{,}675 ScanQA validation questions from 71 scenes at the 9\% reference budget, with one evaluation process per RTX A6000 GPU.
\Cref{tab:eff} averages the per-backbone measurements, including the speedup ratios, over InternVL3-8B, Qwen3-VL-8B, and LLaVA-OneVision-7B.
\ours{} uses the compute-matched configuration with Stage~1 at $\ell_0=\mathrm{round}(d/2)$ and $M'$ input tokens on the applicable backbones; LLaVA-OneVision omits Stage~1 and keeps its original configuration.
Pre-LLM time covers visual encoding, token selection, and projection; time to first token (TTFT) adds the LLM prefill, and total latency adds decoding.
Inputs are resident on the GPU, so image and depth loading are excluded consistently across methods.
Peak memory is the mean per-question peak allocated GPU memory, including model weights.
Accuracy comes from the main evaluation of \cref{tab:adv_main}, while latency and memory come from the profiling runs.

\end{document}